\documentclass[]{spie}  %>>> use for US letter paper
\usepackage[]{graphicx}
\usepackage{caption}
\usepackage{subcaption}
\usepackage{float}
\usepackage{amssymb}
\usepackage{amsmath}
\usepackage{bm}
\usepackage{url}

\usepackage{hyperref}

\newcommand{\vect}[1]{\mathbf{#1}}

\newcommand*\robbar[1]{%
  \hbox{%
    \vbox{%
      \hrule height 0.5pt % The actual bar
      \kern0.25ex%         % Distance between bar and symbol
      \hbox{%
        \kern-0.1em%      % Shortening on the left side
        \ensuremath{#1}%
        \kern-0.1em%      % Shortening on the right side
      }%
    }%
  }%
}

\graphicspath{{figures/}} % NOTE: the last / is important!!

\title{Smooth Extrapolation of Unknown Anatomy via Statistical Shape Models} 

\author{R. B. Grupp\supit{a}, H. Chiang\supit{a}, Y. Otake\supit{a,b}, R. J. Murphy\supit{c,d}, C. R. Gordon\supit{e,f}, M. Armand\supit{c,d},\\and R. H. Taylor\supit{a}
\skiplinehalf
\supit{a}Department of Computer Science, Johns Hopkins University, Baltimore, MD, USA\\
\supit{b}Graduate School of Information Science, Nara Institute of Science and Technology, Nara, Japan\\
\supit{c}Research and Exploratory Development Department, Johns Hopkins University Applied Physics Laboratory, Laurel, MD, USA\\
\supit{d}Department of Mechanical Engineering, Johns Hopkins University, Baltimore, MD, USA\\
\supit{e}Department of Plastic and Reconstructive Surgery, Johns Hopkins University School of Medicine, The Johns Hopkins Hospital, Baltimore, MD, USA\\
\supit{f}Department of Neurosurgery, Johns Hopkins University School of Medicine, The Johns Hopkins Hospital, Baltimore, MD, USA
}

\authorinfo{Send correspondence to R.B. Grupp: grupp@jhu.edu\\
Copyright 2015 Society of Photo-Optical Instrumentation Engineers (SPIE). One print or electronic copy may be made for personal use only. Systematic reproduction and distribution, duplication of any material in this publication for a fee or for commercial purposes, and modification of the contents of the publication are prohibited.\\
Proc. of SPIE Vol. 9415, Medical Imaging 2015: Image-Guided Procedures, Robotic Interventions, and Modeling, 941524\\
DOI: \href{https://www.doi.org/10.1117/12.2081310}{10.1117/12.2081310}}
\begin{document} 
\maketitle 
%%%%%%%%%%%%%%%%%%%%%%%%%%%%%%%%%%%%%%%%%%%%%%%%%%%%%%%%%%%%% 
\begin{abstract}
Several methods to perform extrapolation of unknown anatomy were evaluated.
The primary application is to enhance surgical procedures that may use partial medical images or medical images of incomplete anatomy.
Le Fort-based, face-jaw-teeth transplant is one such procedure.
From CT data of 36 skulls and 21 mandibles separate Statistical Shape Models of the anatomical surfaces were created.
Using the Statistical Shape Models, incomplete surfaces were projected to obtain complete surface estimates.
The surface estimates exhibit non-zero error in regions where the true surface is known; it is desirable to keep the true surface and seamlessly merge the estimated unknown surface.  Existing extrapolation techniques produce non-smooth transitions from the true surface to the estimated surface, resulting in additional error and a less aesthetically pleasing result.
The three extrapolation techniques evaluated were: copying and pasting of the surface estimate (non-smooth baseline), a feathering between the patient surface and surface estimate, and an estimate generated via a Thin Plate Spline trained from displacements between the surface estimate and corresponding vertices of the known patient surface.
Feathering and Thin Plate Spline approaches both yielded smooth transitions.
However, feathering corrupted known vertex values.
Leave-one-out analyses were conducted, with 5\% to 50\% of known anatomy removed from the left-out patient and estimated via the proposed approaches.
The Thin Plate Spline approach yielded smaller errors than the other two approaches, with an average vertex error improvement of 1.46 mm and 1.38 mm for the skull and mandible respectively, over the baseline approach.
\end{abstract}

%>>>> Include a list of keywords after the abstract 


%%%%%%%%%%%%%%%%%%%%%%%%%%%%%%%%%%%%%%%%%%%%%%%%%%%%%%%%%%%%%
\section{INTRODUCTION}
\label{sec:intro}  % \label{} allows reference to this section
\begin{figure}
        \centering
        \includegraphics[width=0.65\textwidth]{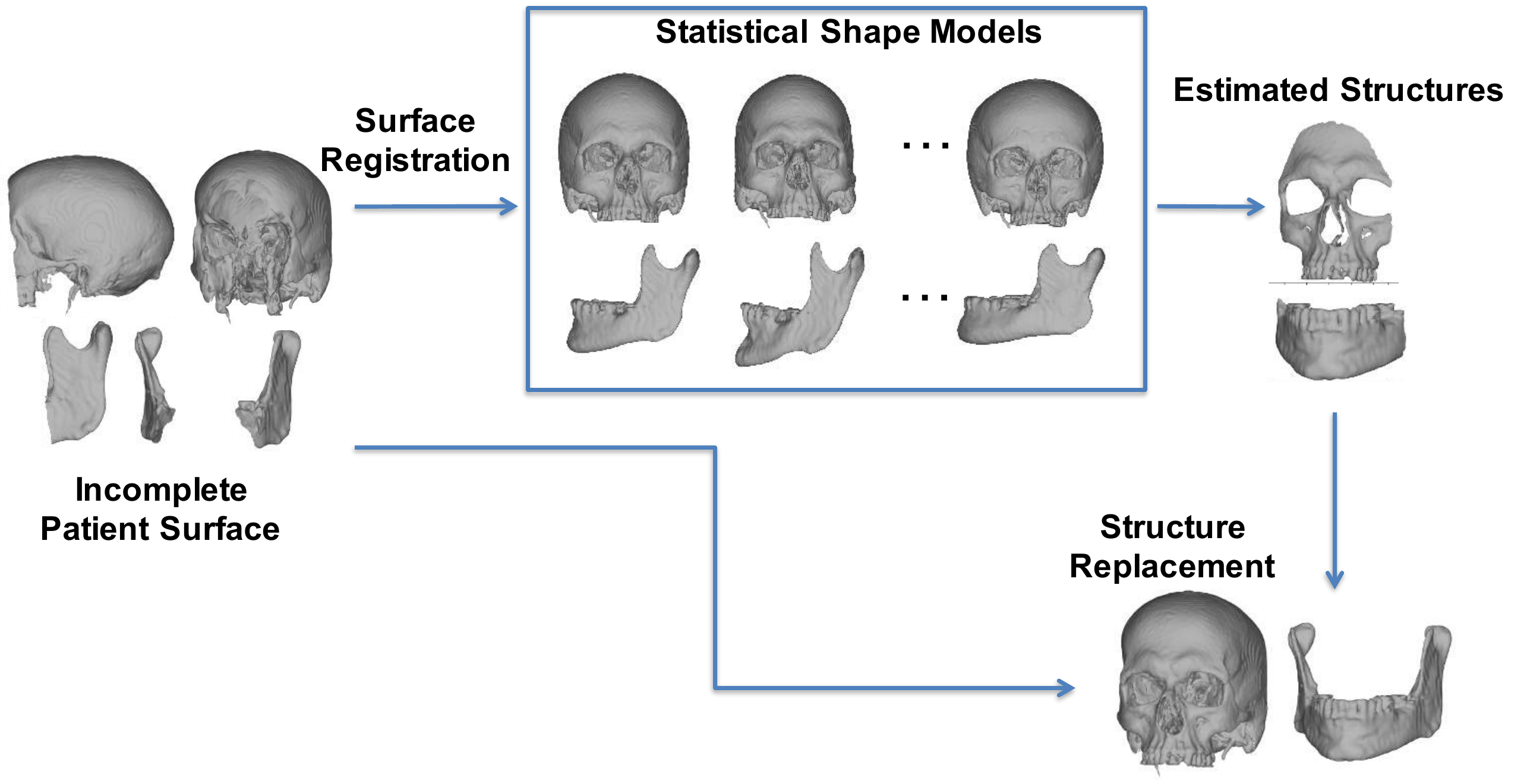}
        \caption{High-level overview of the proposed processing chain.
        	             The contributions of this paper are primarily in the structure replacement component, specifically ensuring smoothness when extrapolating the structure.}
        \label{fig:overview}
\end{figure}

The extrapolation of unknown anatomy is a topic of interest for several medical applications.
For example, consider a patient who has undergone some severe trauma resulting in the loss, or corruption, of skeletal anatomy. In the case of a craniomaxillofacial injury,
%(e.g. shotgun blast), then
Le Fort-based, face-jaw-teeth transplant surgery could be a viable surgical procedure.
Le Fort-based, face-jaw-teeth transplantation aims to transplant an immunocompatible donor's hard and soft craniofacial tissue onto a patient exhibiting some severe trauma or disfigurement\cite{gordon2009world}. 
%Besides aesthetic improvement, the surgery restores lost functionality such as intelligible speech, eating solid food, and smell \cite{siemionow2009near}.
In the case where the planned transplant recipient has no available medical imaging scan prior to trauma, it would be impossible to perform cephalometric measurements on the unknown dento-facial-skeletal regions and to assess their pre-existing dental occlusion.
These measurements may be valuable for use in pre-operative planning and intra-operative navigation in an attempt to optimize positioning of the face-jaw-teeth segment during transplantation (i.e. appearance, airway patency, orbital volumes, etc.) - and to evaluate the anatomical compatibility between donor/recipient related to hybrid occlusion\cite{gordon2012fort,gordon2011osteocutaneous}.
As shown by Chintalapani et al.\@, developmental hip dysplasia (DDH) is another condition that may benefit from anatomical extrapolation\cite{chintalapani2010statisticalspie}.
Pelvic osteotomy is a procedure for correcting hip dysplasia\cite{ganz1988new}. During pelvic osteotomy a fragment of bone, including the acetabular cup, is detached from the pelvis and repositioned to improve the coverage of the femur by the acetabulum. Preoperative CT scans are useful for appropriate three-dimensional geometrical and biomechanical planning of the fragment repositioning\cite{armiger2009three,murphy2014development}.
Most patients undergoing pelvic osteotomy adolescents and females of childbearing age, therefore a partial CT of the pelvis is preferred to minimize radiation dose.
Chintalapani et al.\ utilized a Statistical Shape Model (SSM) of the full pelvis surface to perform anatomical extrapolation for pelvic osteotomy \cite{chintalapani2010statisticalspie}.
%Chintalapani et al.\ utilized a Statistical Shape Model (SSM) of the full pelvis surface to perform anatomical extrapolation for an orthopedic application, Periacetabular Osteotomy\cite{chintalapani2010statisticalspie,ganz1988new}.
This extrapolation technique typically results in a non-smooth transition from the patient's true anatomical region
(acetabulum)
into the extrapolated region
(ilium)
\cite{chintalapani2010statistical}.
This non-smooth transition introduces additional error and potentially degrades any associated measurements across regions.  We propose two methods for smoothly extrapolating unknown anatomy and provide comparisons to the baseline approach identified by Chintalapani et al.  One method consists of ``feathering'' the surface values in a region of overlap between the patient surface and SSM surface. The other method utilizes a Thin Plate Spline (TPS), trained from surface displacement values in an overlap region, to displace the surface values of the SSM surface. Figure \ref{fig:overview} shows a high level overview of the processing involved.
%%%%%%%%%%%%%%%%%%%%%%%%%%%%%%%%%
%%%%%%%%%%%%%%%%%%%%%%%%%%%%%%%%%
%%%%%%%%%%%%%%%%%%%%%%%%%%%%%%%%%%%%%%%%%%%%%%%%%%%%%%%%%%%%%
\section{METHODS}
\label{sec:methods}  % \label{} allows reference to this section
\subsection{STATISTICAL SHAPE MODEL}
Separate SSMs of the mandible and craniomaxillofacial skeleton excluding mandible (also referred to as skull) were created from CT data provided by The Cancer Imaging Archive\cite{website:tcia-qin-head-neck,website:tcia-head-neck-cetuximab}.
Separating the mandible from the rest of the craniomaxillofacial skeleton allows for characterization of the mandible independent of its pose relative to the remainder of the skull.
This initial dataset consisted of 59 head CT images.

Prior to any processing, data was resampled to 1 mm isotropic voxels using a B-Spline interpolator of order 3.
Each CT was cropped to contain the entire head volume with an inferior bound at C5.
We implementated the pipeline proposed by Chintalapani et al. to create each SSM\cite{chintalapani2007statistical}.
A template CT was chosen, after which every other CT was registered to the template using a deformable diffeomorphic algorithm \cite{Avants20112033}.
Each template was segmented and a surface mesh created.
Basic thresholding was performed and followed by manual ``touch-up'' to segment the bone structures.
The Marching Cubes\cite{lorensen1987marching} algorithm was performed on each template label map to generate an initial isosurface, followed by a $\textrm{sinc}$-based smoothing and quadric decimation\cite{garland1997surface} to obtain the final surface mesh.
In order to perform statistical analysis, a collection of ``topologically consistent'' meshes is required; meaning the number of vertices and triangles are constant over all meshes, the vertices are homologous between subjects, and the triangular structure is constant.
Using displacement fields from the deformable image registration, each template surface mesh was warped to create topologically consistent meshes for every other subject; tri-linear interpolation was used to weight displacement vectors surrounding each sub-voxel valued mesh vertex.
Each warped mesh was checked manually for abnormal registration results.
This resulted in 36 valid skull surfaces and 21 valid mandible surfaces.
Each skull mesh consisted of 255,340 vertices and 511,863 triangles and each mandible mesh consisted of 34,764 vertices and 69,578 triangles.

Each set of topologically consistent meshes were aligned via a Generalized Procrustes Analysis\cite{gower1975generalized}, accounting for rigid pose and scale.
Each aligned mesh's vertices were stacked into a large column vector and Principle Component Analysis was performed over these vectors to define the SSM.  An instance, $V$, of the SSM is represented by a linear combination of principal component vectors ($v_1, \dots, v_N$) and the mean shape ($v_0$) as shown in \eqref{eq:ssm_instance}. This is the classical Point Distribution Model\cite{Cootes199538}.
Given a patient mesh, $V_P$,  that is topologically consistent with the SSM, the  best approximating SSM instance, $\widetilde{V}_P$, is represented by the projection of $V_P$ onto the SSM as shown in \eqref{eq:ssm_proj}.
$\mathcal{T}$ represents the similarity transformation of the patient mesh to the SSM mean, $v_0$.
The projection operator in \eqref{eq:ssm_proj} shall be denoted as $\mathcal{P}$.
As described by Chintalapani et al., a leave-one-out analysis was conducted to analyze the generalization capability of each SSM\cite{chintalapani2007statistical}.
\begin{equation}\label{eq:ssm_instance}
  V = v_0 + \sum_{i = 1}^N \lambda_i v_i
\end{equation}
\begin{equation}\label{eq:ssm_proj}
  \widetilde{V}_P = \mathcal{P}\left(V_P\right) = \mathcal{T}^{-1}\left(v_0 + \sum_{i=1}^N \left( \mathcal{T}\left(V_P\right) - v_0 \right)^T v_i\right)
\end{equation} 
\subsection{EXTRAPOLATION}
\begin{figure}
        \centering
        \includegraphics[width=0.75\textwidth]{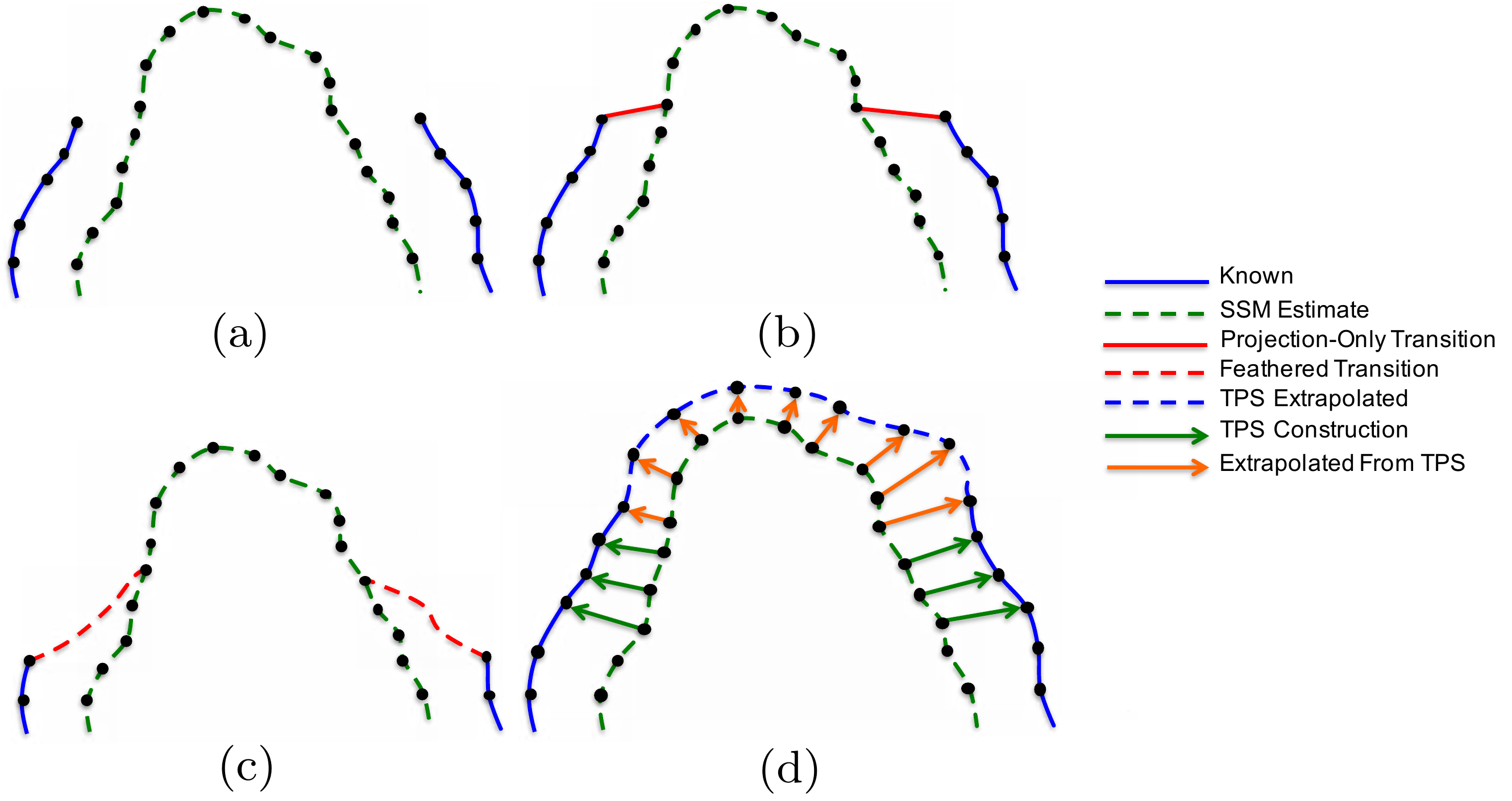}
        \caption{Toy example of the extrapolation strategies. (a) Incomplete patient (solid blue) and a SSM-derived estimate of the entire surface (dashed green). (b) Projection Only approach with non-smooth transition (solid red). (c) Projection Plus Feathering with modified overlap region (dashed red). (d) Projection Plus TPS with vertex pairs used to construct the TPS (solid green arrows), extrapolated vertices (orange arrows), and the extrapolated surface (dashed blue)} \label{fig:extrap_toy}
\end{figure}

Suppose the patient mesh, $V_P$, is partitioned into two regions, one representing the ``known'' vertex values and the other representing the ``unknown,'' or missing, vertex values. $V_P$ may therefore be partitioned as $(V_P^\text{known}, V_P^\text{unknown})^T$.  The extrapolation process attempts to estimate the values of $V_P^\text{unknown}$ given $V_P^\text{known}$ and a SSM: $\{ v_0, v_1, \dots, v_N \}$.  Let $V_P^0$ represent the full-length vector $V_P$ with $V_P^\text{unknown}$ set to zero, e.\@g.\ $V_P^0 = (V_P^\text{known}, \vect{0})^T$. By projecting $V_P^0$ onto the SSM, an estimate of the entire anatomy, $\widetilde{V}_P = \mathcal{P}(V_P^0)$, is obtained.
%; shown in \eqref{eq:ssm_proj_extrap}.
%\begin{equation}\label{eq:ssm_proj_extrap}
%	\widetilde{V}_P = \mathcal{T}^{-1}\left(v_0 + \sum_{i=1}^N \left( \mathcal{T}\left(V_P^0\right) - v_0 \right)^T v_i\right)
%\end{equation}
The partitioning of $V_P$ may be applied to $\widetilde{V}_P$, yielding $\widetilde{V}_P = (\widetilde{V}_P^\text{known}, \widetilde{V}_P^\text{unknown})^T$.
The extrapolation method described by Chintalapani et al.\cite{chintalapani2010statisticalspie} is equivalent to replacing $V_P^\text{unknown}$ with $\widetilde{V}_P^\text{unknown}$, resulting in the output vertices: $(V_P^\text{known}, \widetilde{V}_P^\text{unknown})^T$. This is analogous to ``cutting'' the missing vertex values from the SSM instance and ``pasting'' them into the true patient surface. As identified by Chintalapani, this process results in a non-smooth transition from the known region into the unknown region\cite{chintalapani2010statistical}.  We shall refer to this approach as Projection Only (PO). Figure \ref{fig:extrap_toy}b highlights the extrapolation with a simple example.

The remaining extrapolation techniques require a region of overlap between the patient and SSM instance meshes. This region is computed by traversing along triangle edges in the known region, and originating from the boundary between known and unknown regions.  Once a specific depth, $d$, has been reached the traversal is terminated.

The ``feathering'' approach is derived from methods used in image panorama creation to ``stitch'' consecutive frames together\cite{szeliski2006image}. Our implementation of feathering computes weighted combinations of vertex values from $V_P^\text{known}$ and $\widetilde{V}_P^\text{known}$ confined to the region of overlap between the SSM estimate and known patient.
Let $q$ denote a vertex in the true patient mesh that is located in the overlap region, let $r$ denote the corresponding vertex in the SSM estimate, and let $n$ be the minimum number of edge traversals to reach $q$ and $r$ from the boundary.  The feathered vertex value, $\widetilde{q}$, is computed as in \eqref{eq:feather} and replaces the value of $q$.
All vertices in the unknown region are copied and pasted just as in the Projection Only approach.
This method has the undesirable property of corrupting the patient's true vertex values in the overlap region.
We shall refer to this approach as Projection Plus Feathering (P+F). A graphical example of this approach is shown in Figure \ref{fig:extrap_toy}c.
\begin{equation}\label{eq:feather}
  \widetilde{q} = \frac{d - n}{d} r + \frac{n}{d} q
\end{equation}

The TPS approach uses the displacements from vertices in the SSM estimate to the corresponding vertices in the patient mesh.  Using these displacements in the overlap region, the TPS is constructed as described by Bookstein\cite{Bookstein24792}. Every vertex value of the SSM estimate in the unknown region is input to the TPS and the output is used to set the corresponding vertex in the patient's unknown region.
The TPS implicitly computes the global affine transformation components to align the unknown region of the SSM estimate with the boundary of the patient's known region. The smoothness properties of TPS further ensure that transition between the known and unknown regions is smooth.
We shall refer to this approach as Projection Plus TPS (P+TPS). Figure \ref{fig:extrap_toy}d depicts the intuition behind the TPS approach.
P+TPS is much more computationally complex than P+F, since computation of the TPS weights involves the solution of three dense systems of equations, each of dimension $(N+4) \times (N+4)$ ($N$ is the number of vertices in the overlap region). Therefore, P+TPS is typically run with much smaller overlap regions than P+F. 
In order to solve for TPS coefficients, we use a C++ implementation of the QR decomposition with column pivoting using Householder Reflections\cite{eigenweb}. The QR implementation is not multi-threaded, however we solve the three systems in parallel. Once the TPS weights are computed, the extrapolated points may be computed fully in parallel.

For evaluation, leave-one-out tests similar in nature to those performed on the SSMs are used, however instead of varying the number of modes used during each iteration, the amount of known anatomy is varied.
For each part of anatomy, the set of meshes are aligned and the mean shape is computed. A bounding box is computed about the mean shape and a percentage of anatomy is removed in increments of 5\%.
Cropping percentages of 5\% through 50\% were used with the region beginning at the anterior extent and moving along the sagittal plane; this corresponds to cropping the ``front of the face.''
Examples of this cropping is shown in figure \ref{fig:skull_extrap_examples} (20\% cropped) and figure \ref{fig:mand_extrap_examples} (50\% cropped).
For each leave-out iteration, the left-out mesh is excluded and a SSM is created from the remaining meshes.
For each cropping specification, the left-out mesh is cropped and the remaining vertices are projected onto the corresponding vertices of the SSM to obtain mode weights (as previously defined).
An estimate of the entire anatomy is instantiated via the mode weights and each of the extrapolation methods are executed.
When projecting onto the SSM, all available modes are used.
% TODO: talk about error statistics
Error statistics are computed in the unknown regions for PO and P+TPS, and over the union of overlap and unknown regions for P+F, since feathering introduces errors into the overlap region.
%The error statistics measured are a Modified Hausdorff Distance, Hausdorff Distance, and root mean square vertex error.
%TODO: provide formulas
%Given two meshes, $\mathcal{M}_1$ and $\mathcal{M}_2$, let $\mathcal{O}_1$ denote the set vertices from $\mathcal{M}_1$ lying in the overlap region.
%%%%%%%%%%%%%%%%%%%%%%%%%%%%%%%%%%%%%%%%%%%%%%%%%%
%%%%%%%%%%%%%%%%%%%%%%%%%%%%%%%%%%%%%%%%%%%%%%%%%%%%%%%%%%%%%
\section{RESULTS}
\label{sec:results}  % \label{} allows reference to this section
\begin{figure}
        \centering
        \begin{subfigure}[b]{0.2\textwidth}
                \includegraphics[width=\textwidth, clip=true, trim=145 60 125 35]{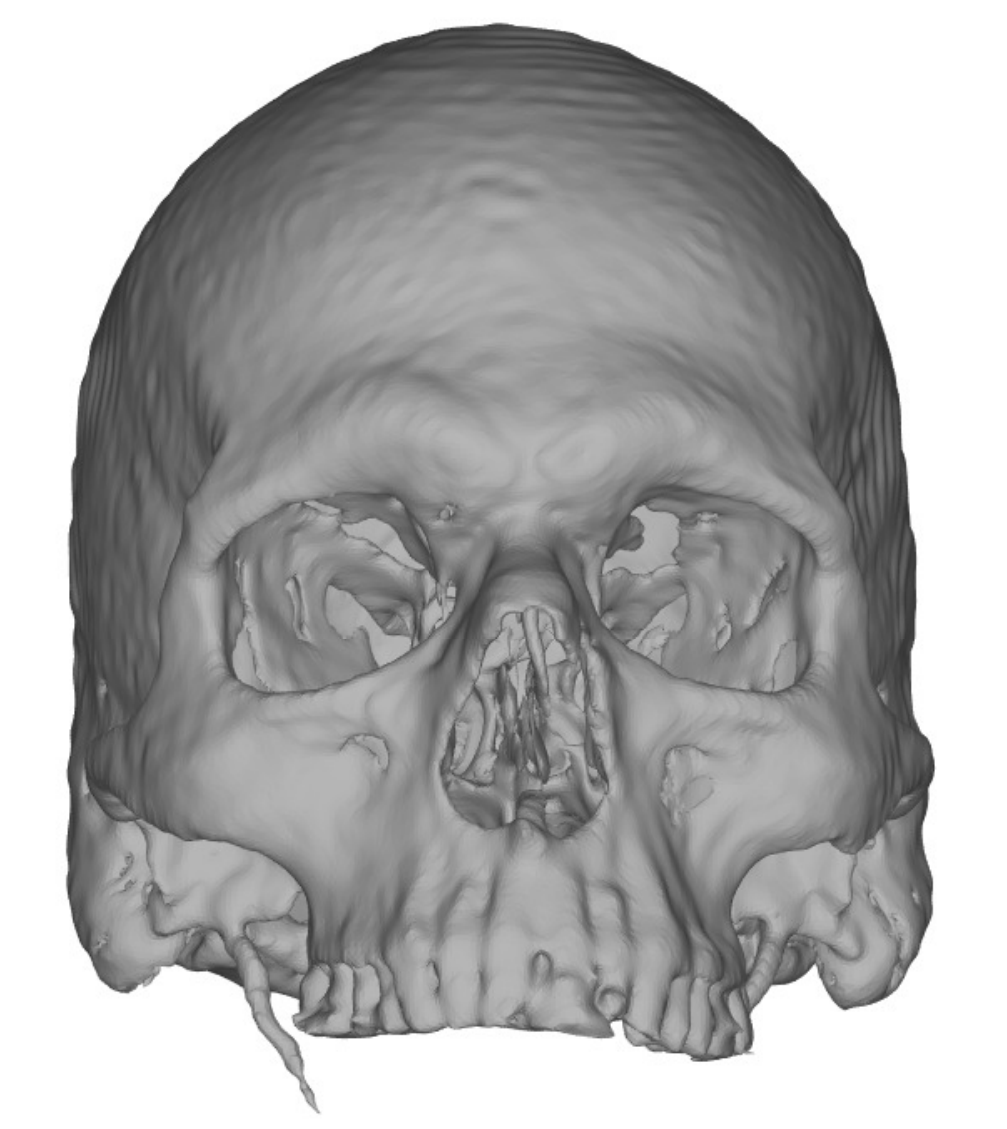}
                \caption{}
                \label{fig:skull_mode1_minus3}
        \end{subfigure}
        \begin{subfigure}[b]{0.2\textwidth}
                \includegraphics[width=\textwidth, clip=true, trim=145 60 125 35]{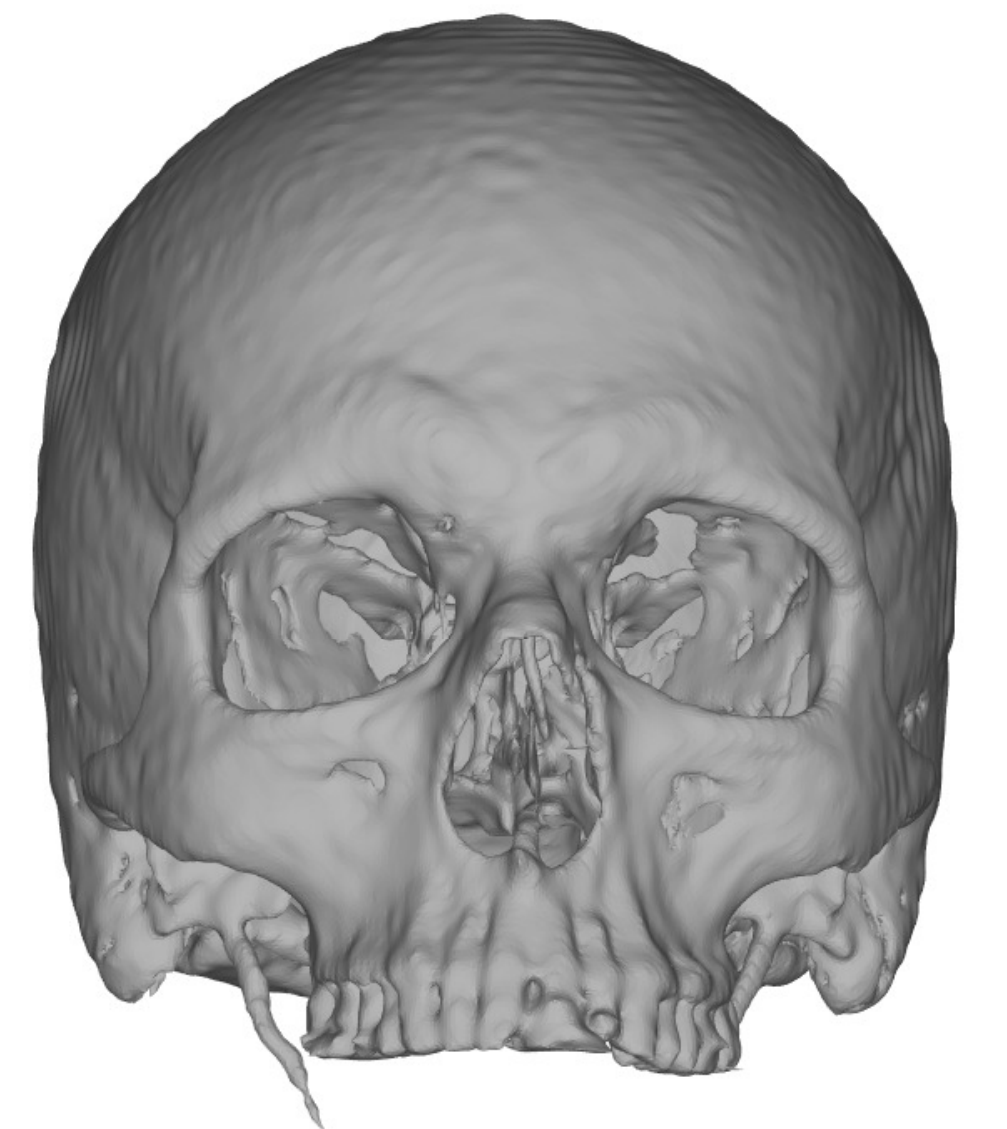}
                \caption{}
                \label{fig:skull_mean}
        \end{subfigure}
        \begin{subfigure}[b]{0.2\textwidth}
                \includegraphics[width=\textwidth, clip=true, trim=145 60 125 35]{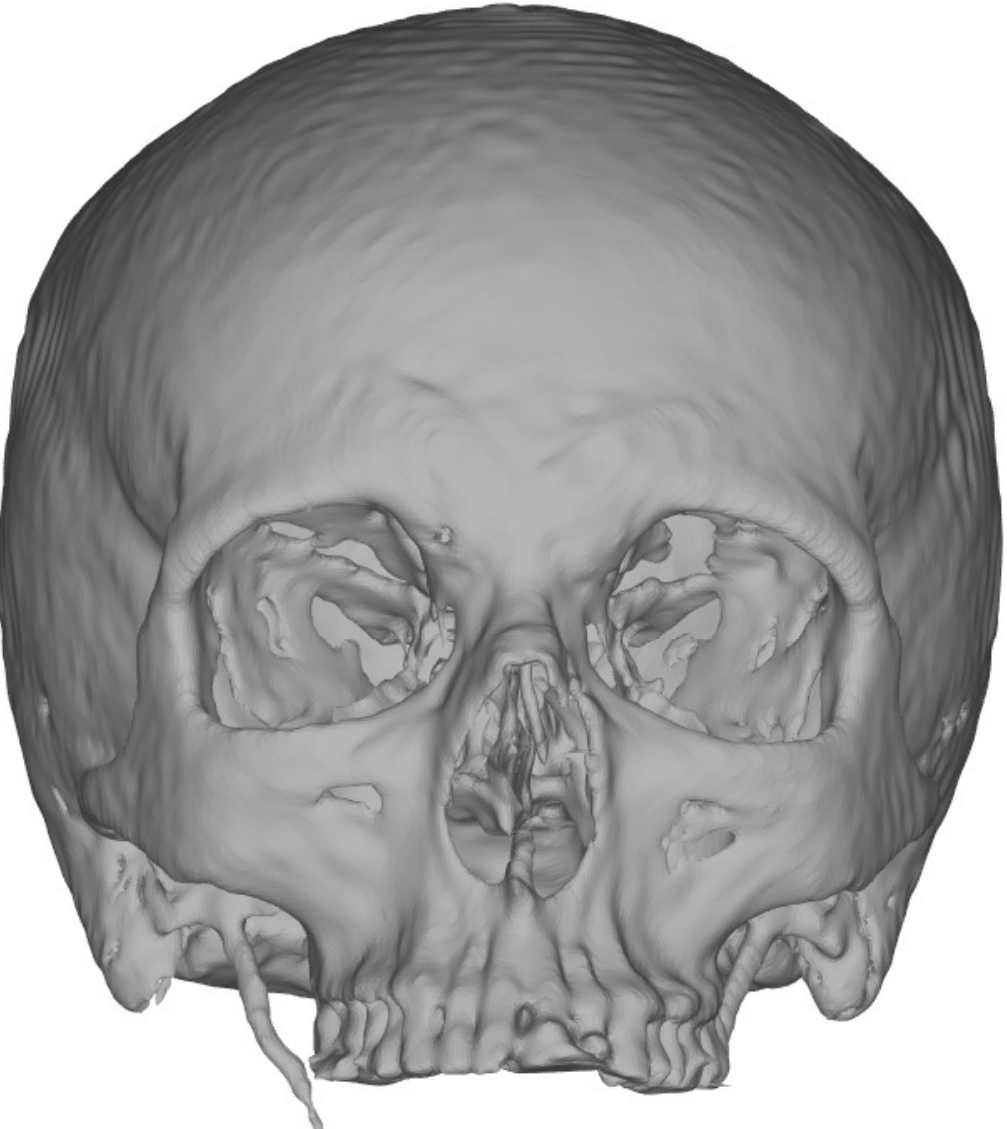}
                \caption{}
                \label{fig:skull_mode1_plus3}
        \end{subfigure}\\
        \begin{subfigure}[b]{0.2\textwidth}
                \includegraphics[width=\textwidth, clip=true, trim=90 45 70 35]{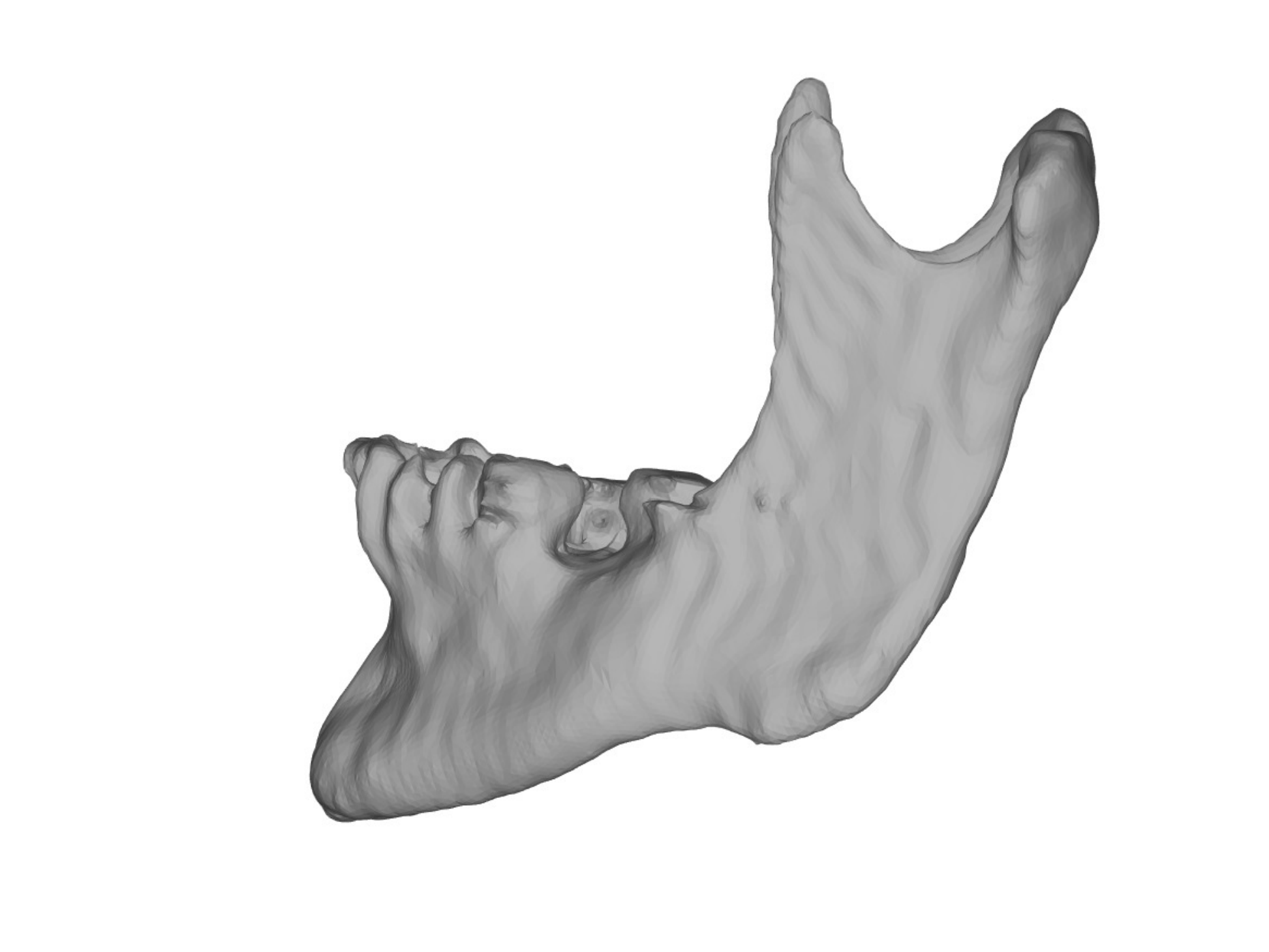}
                \caption{}
                \label{fig:mand_mode1_minus3}
        \end{subfigure}
        \begin{subfigure}[b]{0.2\textwidth}
                \includegraphics[width=\textwidth, clip=true, trim=90 45 70 35]{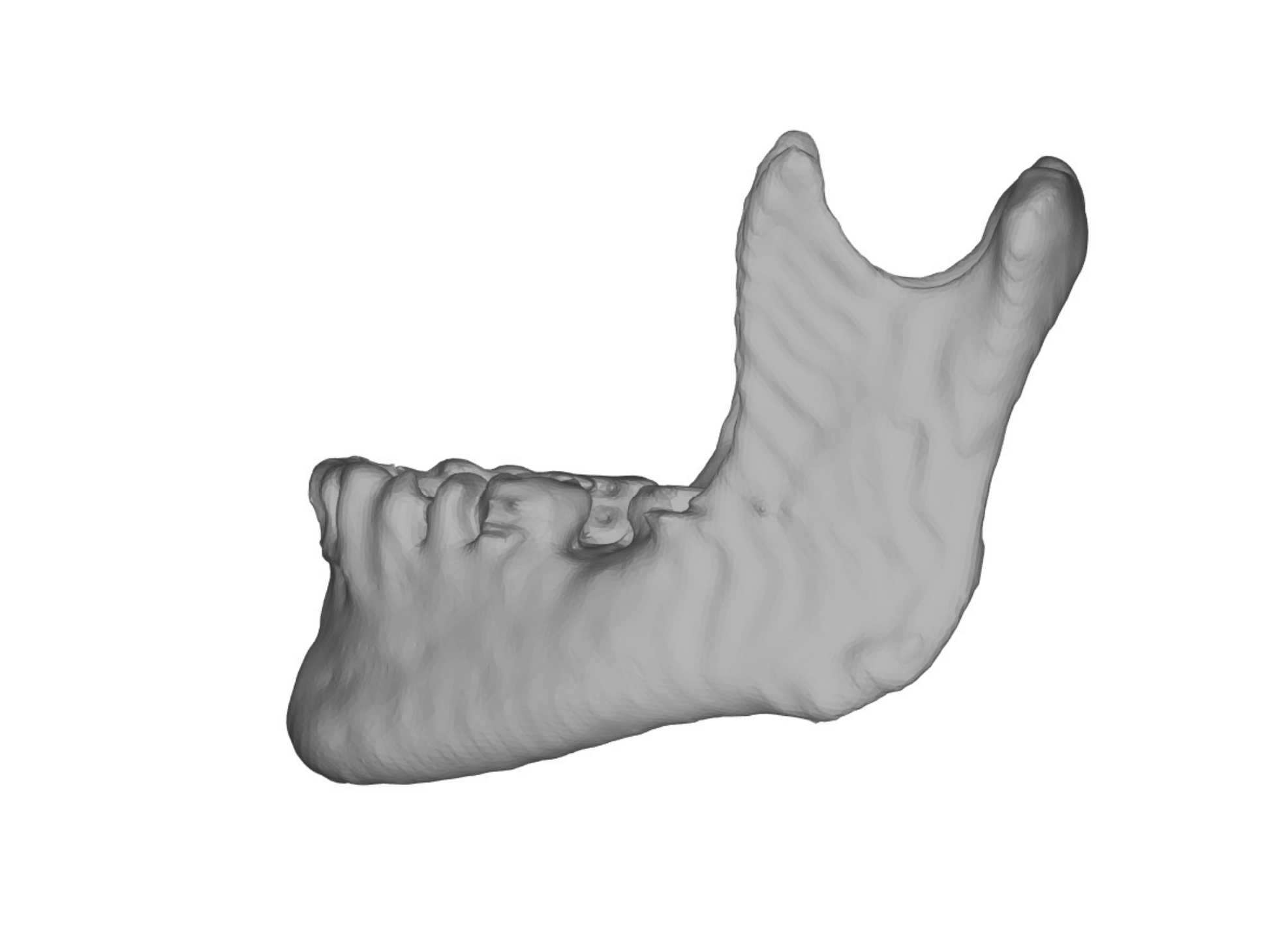}
                \caption{}
                \label{fig:mand_mean}
        \end{subfigure}
        \begin{subfigure}[b]{0.2\textwidth}
                \includegraphics[width=\textwidth, clip=true, trim=90 45 70 35]{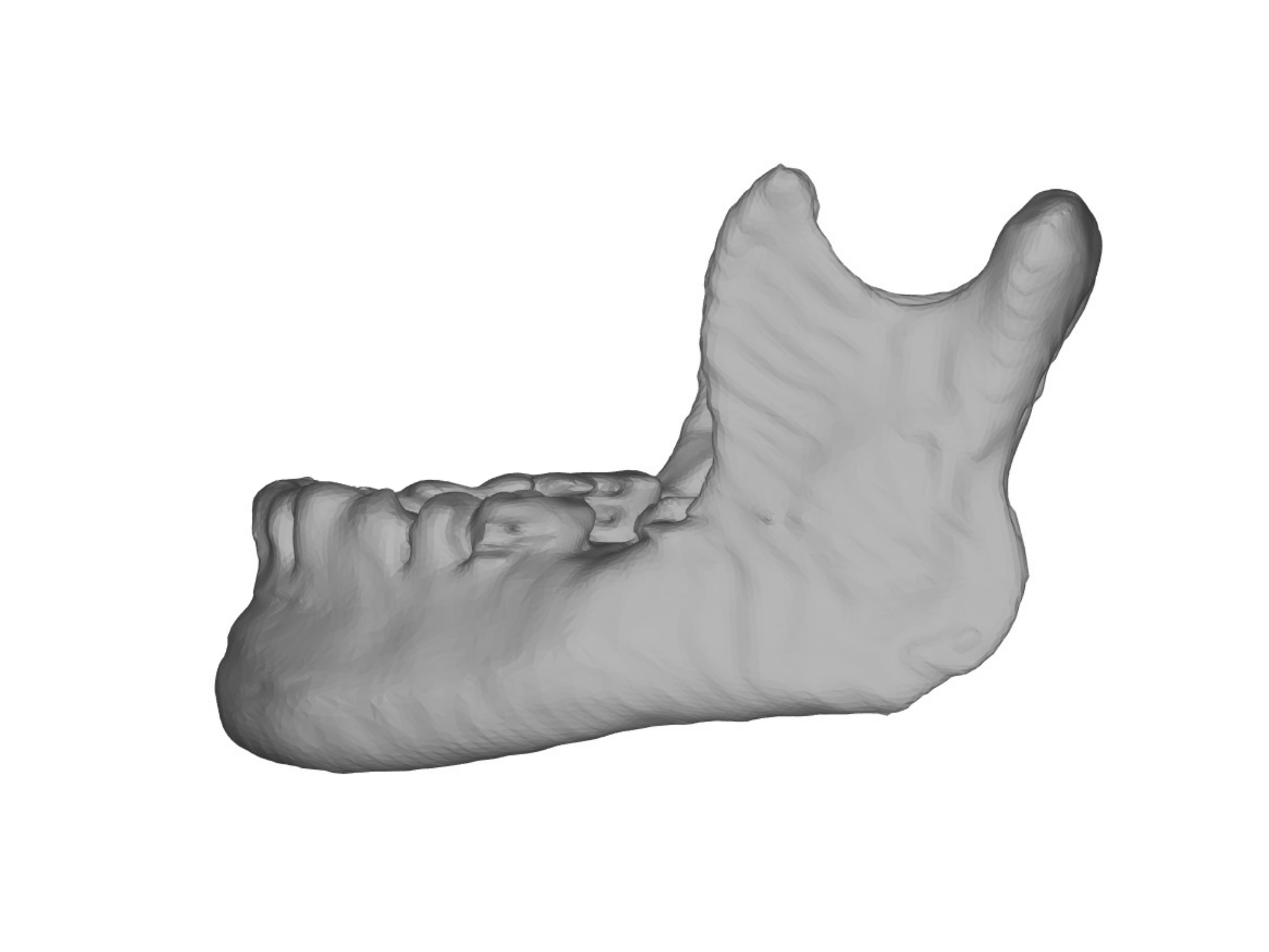}
                \caption{}
                \label{fig:mand_mode1_plus3}
        \end{subfigure}
        \caption{First principal modes of each surface $\pm 3$ standard deviations.
        		     (a) Skull $v_0 - 3 \sigma v_1$.
		     (b) Skull mean shape $v_0$.
		     (c) Skull $v_0 + 3 \sigma v_1$.
		     (d) Mandible $v_0 - 3 \sigma v_1$.
		     (e) Mandible mean shape $v_0$.
		     (f) Mandible $v_0 + 3 \sigma v_1$.
		     } \label{fig:mode_shapes}
\end{figure}

\begin{figure}
        \centering
        \begin{subfigure}[b]{0.42\textwidth}
                \includegraphics[width=\textwidth]{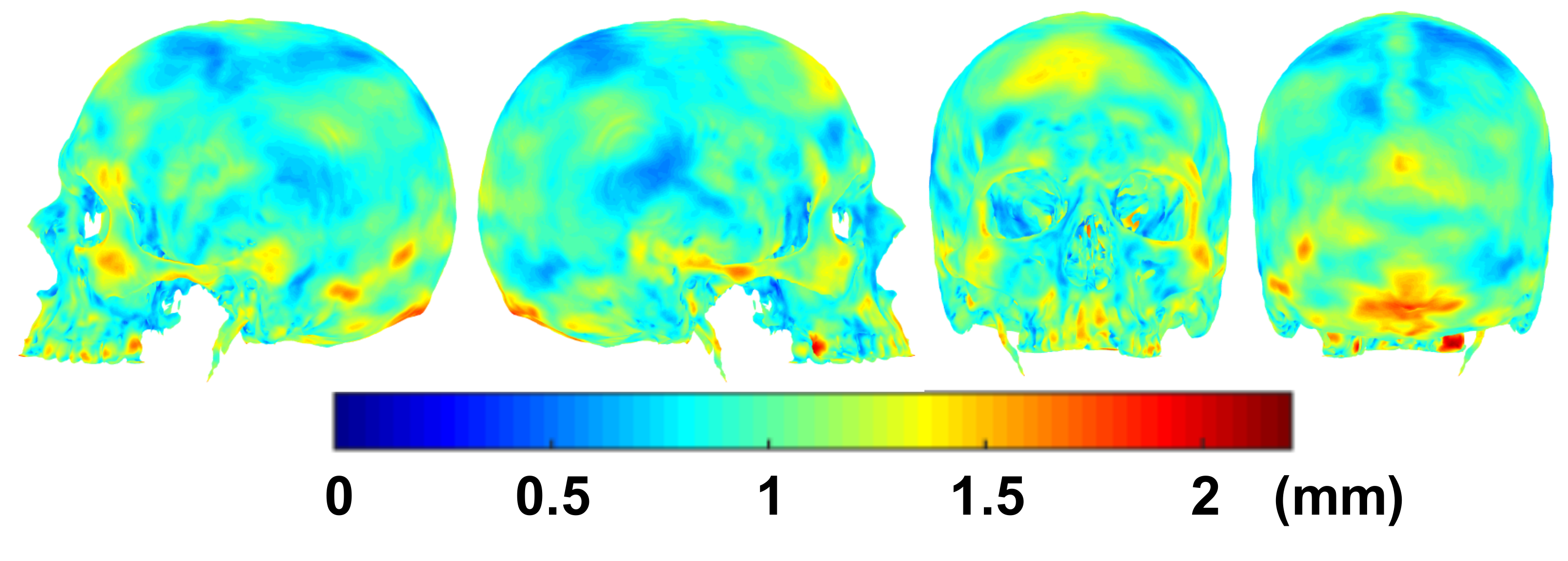}
                \caption{}
                \label{fig:skull_atlas_heat_map}
        \end{subfigure}
        \begin{subfigure}[b]{0.47\textwidth}
                \includegraphics[width=\textwidth]{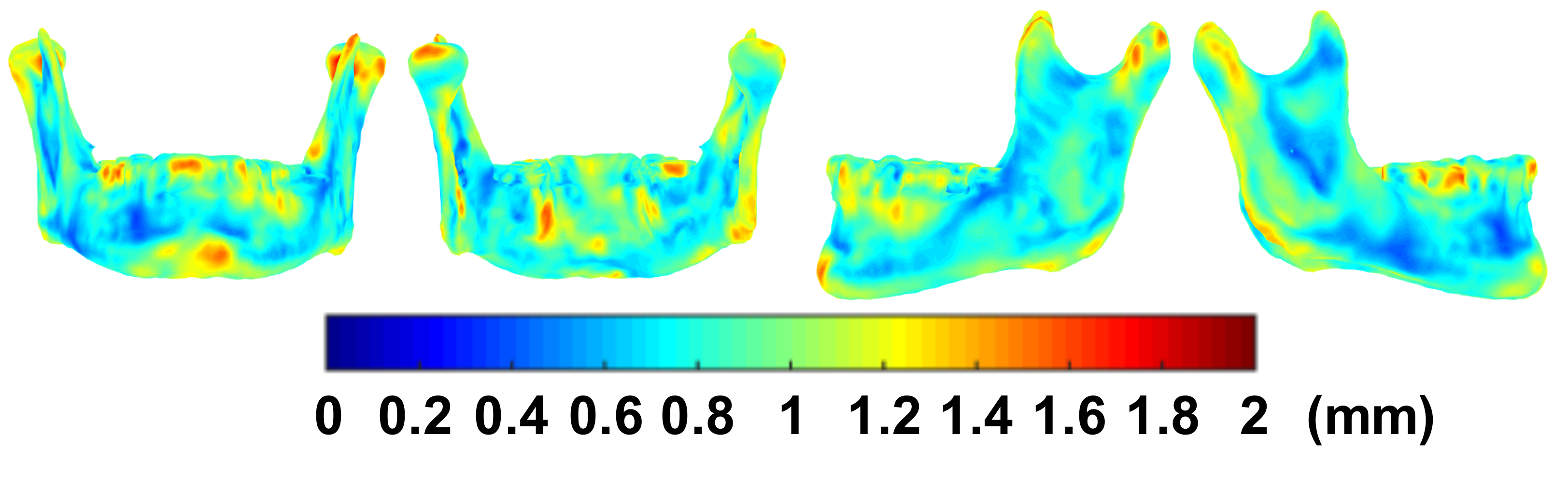}
                \caption{}
                \label{fig:mand_atlas_heat_map}
        \end{subfigure}
        \caption{Heat maps of mean surface deviations from a left-out surface to the SSM instance.
        The distance between each true vertex and the closest point on the SSM estimate is computed and averaged over each leave-out trial. The underlying surfaces are the mean shapes of all training data.
        This shows that the SSMs are able to reproduce a surface with approximately 1.0 to 1.5 mm of surface error.
        Localized areas of larger error may be identified, such as the teeth, chin, and base of skull.} \label{fig:atlas_heat_maps}
\end{figure}

Using all available meshes, SSMs were created to understand the principal modes of variation for the skull and mandible. Figure \ref{fig:mode_shapes} shows the mean shape and one standard deviation of the first mode for each bone.
As a result of leave-out testing and using all SSM modes, the skull SSM achieved a root mean square (RMS) surface deviation of 1.25 mm
(std.\@ dev.\@ 0.02 mm),
a maximum surface deviation of 6.22 mm (std.\@ dev.\@ 1.38 mm),
and a RMS vertex error of 1.95 mm
(std.\@ dev.\@ 0.06 mm).
Using all SSM modes, the mandible SSM achieved a RMS surface deviation of 1.08 mm
(std.\@ dev.\@ 0.02 mm),
a maximum surface deviation of 4.61 mm (std.\@ dev.\@ 0.83 mm),
and a RMS vertex error of 1.72 mm
(std.\@ dev.\@ 0.08 mm).
The distribution of surface errors for each SSM is shown in Figure \ref{fig:atlas_heat_maps}.
%%%%%%%%%%%%%%%%%%%%%%%%%%%%%%%%%%%%%%%%%%%%%%%%%%
%%%%%%%%%%%%%%%%%%%%%%%%%%%%%%%%%%%%%%%%%%%%%%%%%%%%%%%%%%%%%
%%%%%%%%%%%%%%%%%%%%%%%%%%%%%%%%%%%%%%%%%%%%%%%%%%
%%%%%%%%%%%%%%%%%%%%%%%%%%%%%%%%%%%%%%%%%%%%%%%%%%
%%%%%%%%%%%%%%%%%%%%%%%%%%%%%%%%%%%%%%%%%%%%%%%%%%
\begin{figure}
        \centering
        \begin{subfigure}[b]{0.3\textwidth}
                \includegraphics[width=\textwidth, clip=true, trim=170 75 120 60]{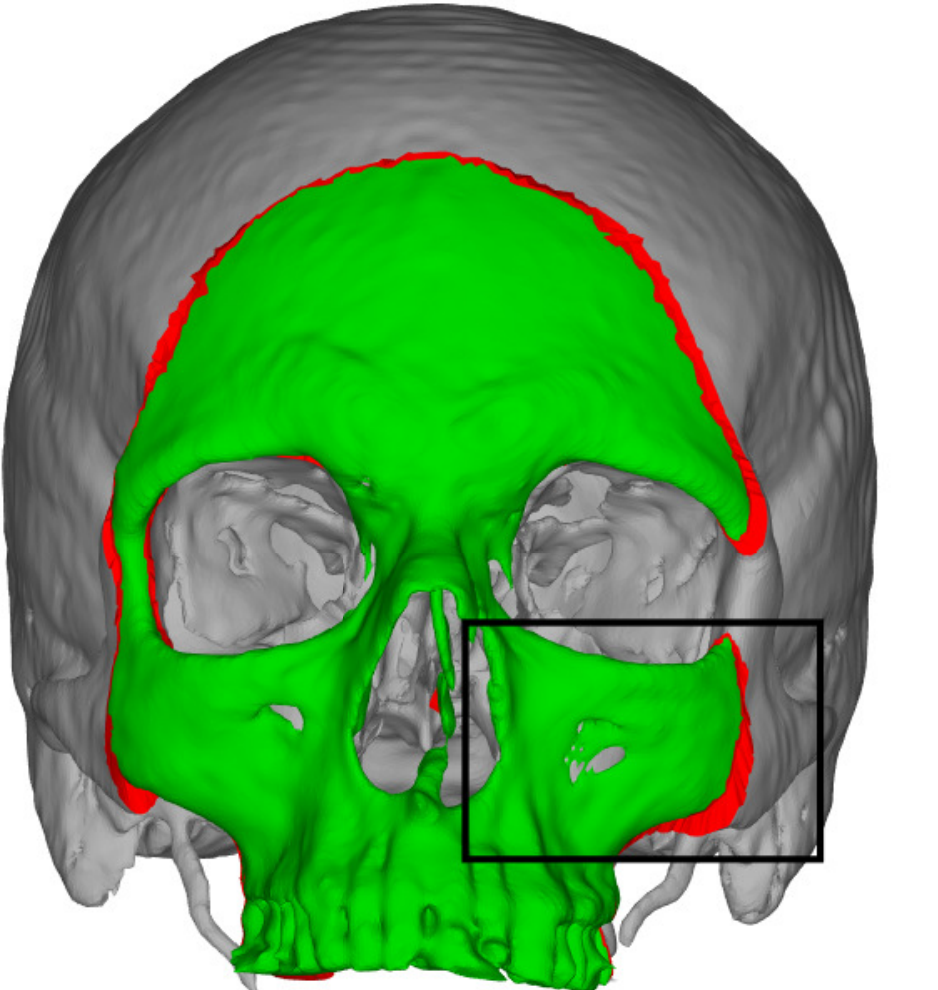}
                \caption{}
                \label{fig:skull_0522c0014_01_def_03_proj_only}
        \end{subfigure}
        \begin{subfigure}[b]{0.3\textwidth}
                \includegraphics[width=\textwidth, clip=true, trim=170 75 120 60]{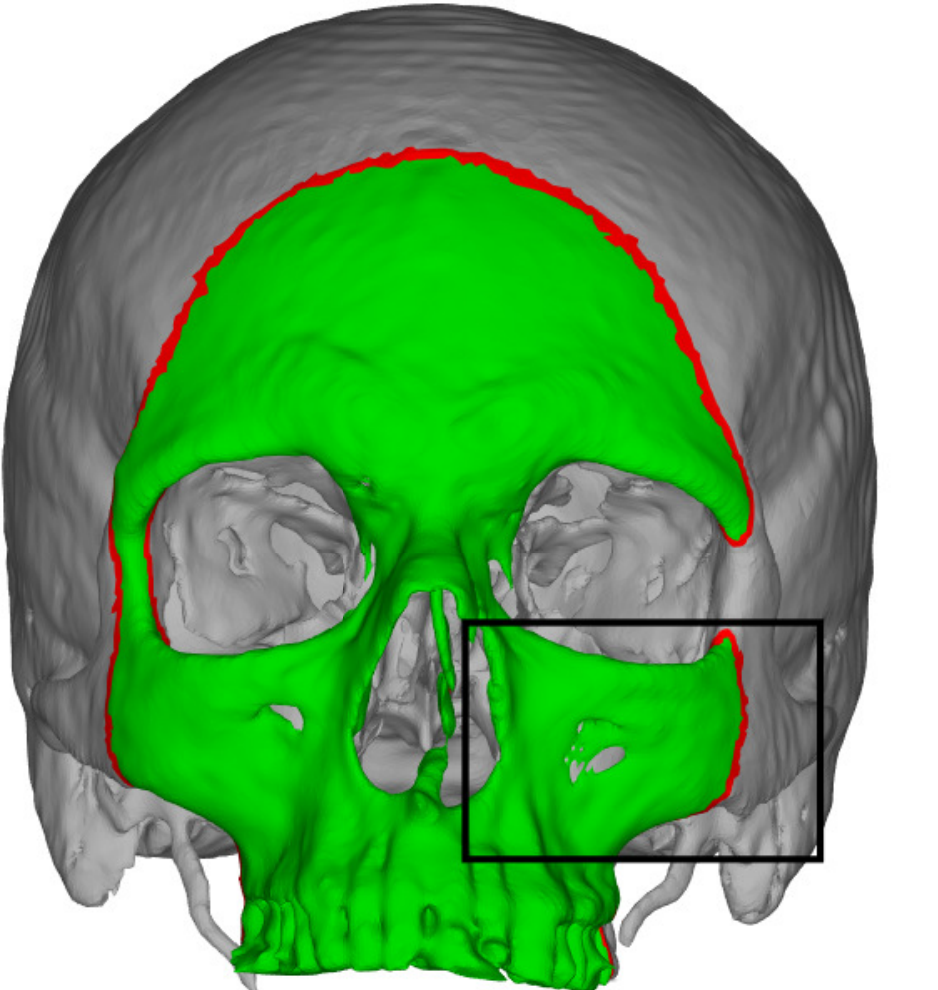}
                \caption{}
                \label{fig:skull_0522c0014_01_def_03_proj_feather}
        \end{subfigure}
        \begin{subfigure}[b]{0.3\textwidth}
                \includegraphics[width=\textwidth, clip=true, trim=170 75 120 60]{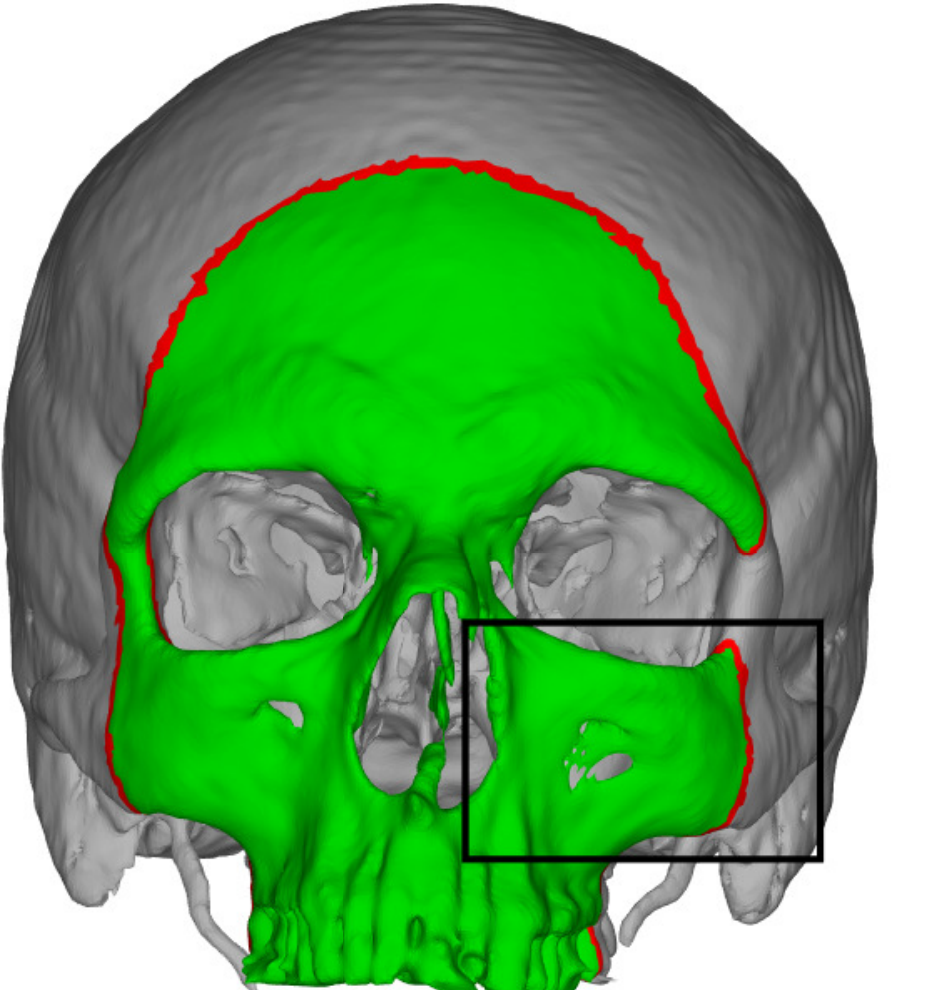}
                \caption{}
                \label{fig:skull_0522c0014_01_def_03_proj_tps}
        \end{subfigure}\\
        \begin{subfigure}[b]{0.3\textwidth}
                \includegraphics[width=\textwidth, clip=true, trim=70 90 50 50]{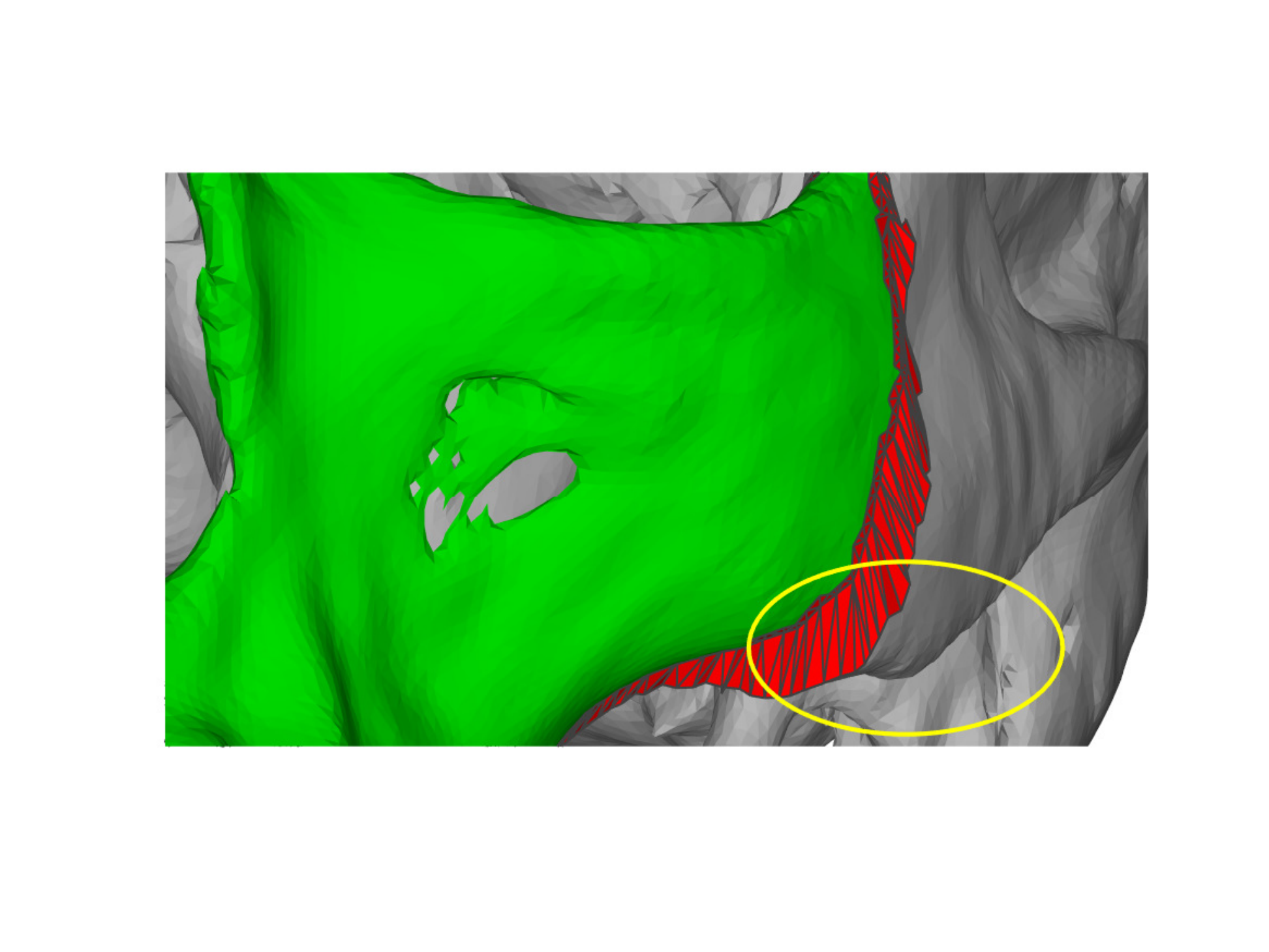}
                \caption{}
                \label{fig:skull_0522c0014_01_def_03_proj_only_zoom}
        \end{subfigure}
        \begin{subfigure}[b]{0.3\textwidth}
                \includegraphics[width=\textwidth, clip=true, trim=70 90 50 50]{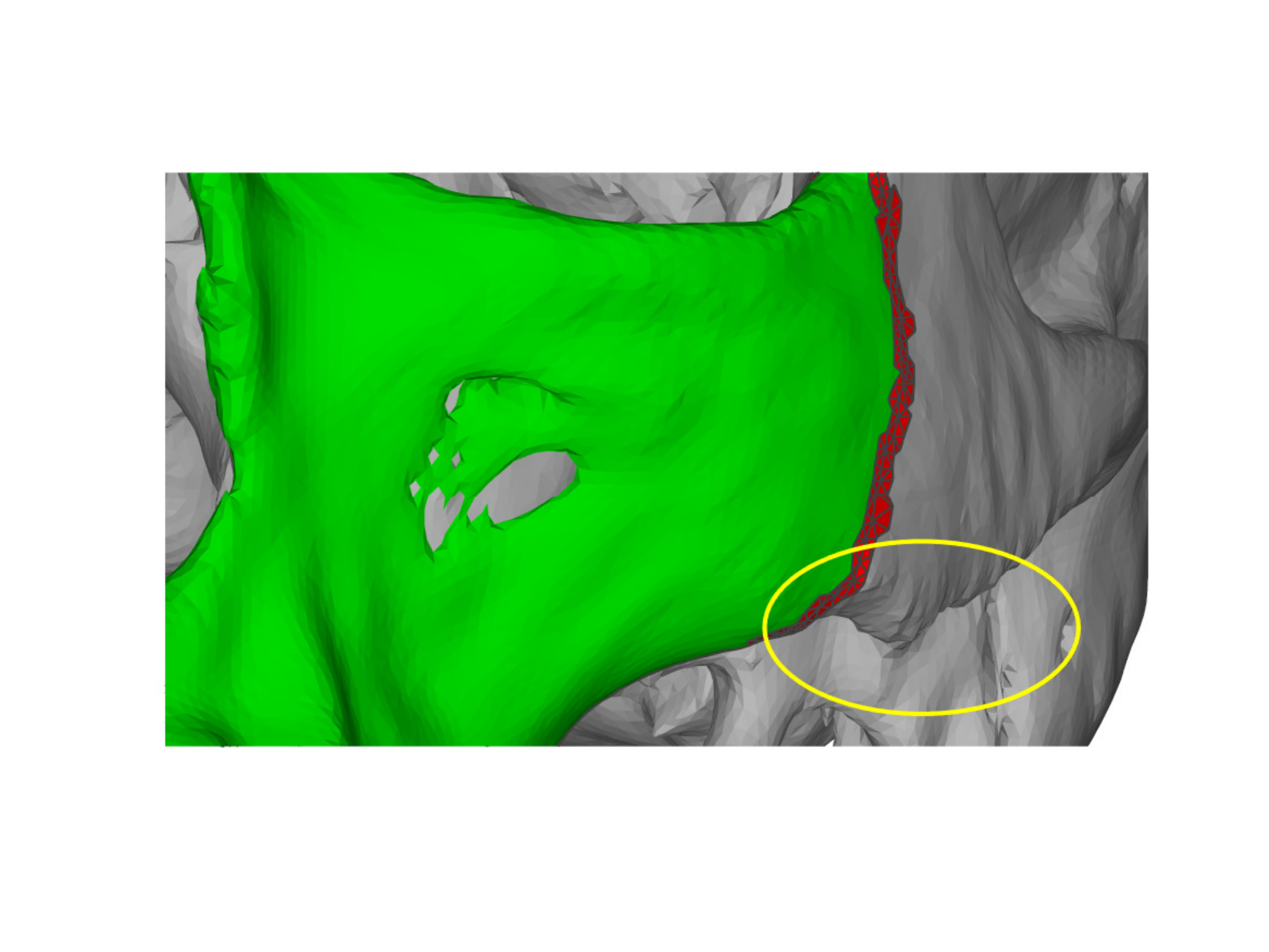}
                \caption{}
                \label{fig:skull_0522c0014_01_def_03_proj_feather_zoom}
        \end{subfigure}
        \begin{subfigure}[b]{0.3\textwidth}
                \includegraphics[width=\textwidth, clip=true, trim=70 90 50 50]{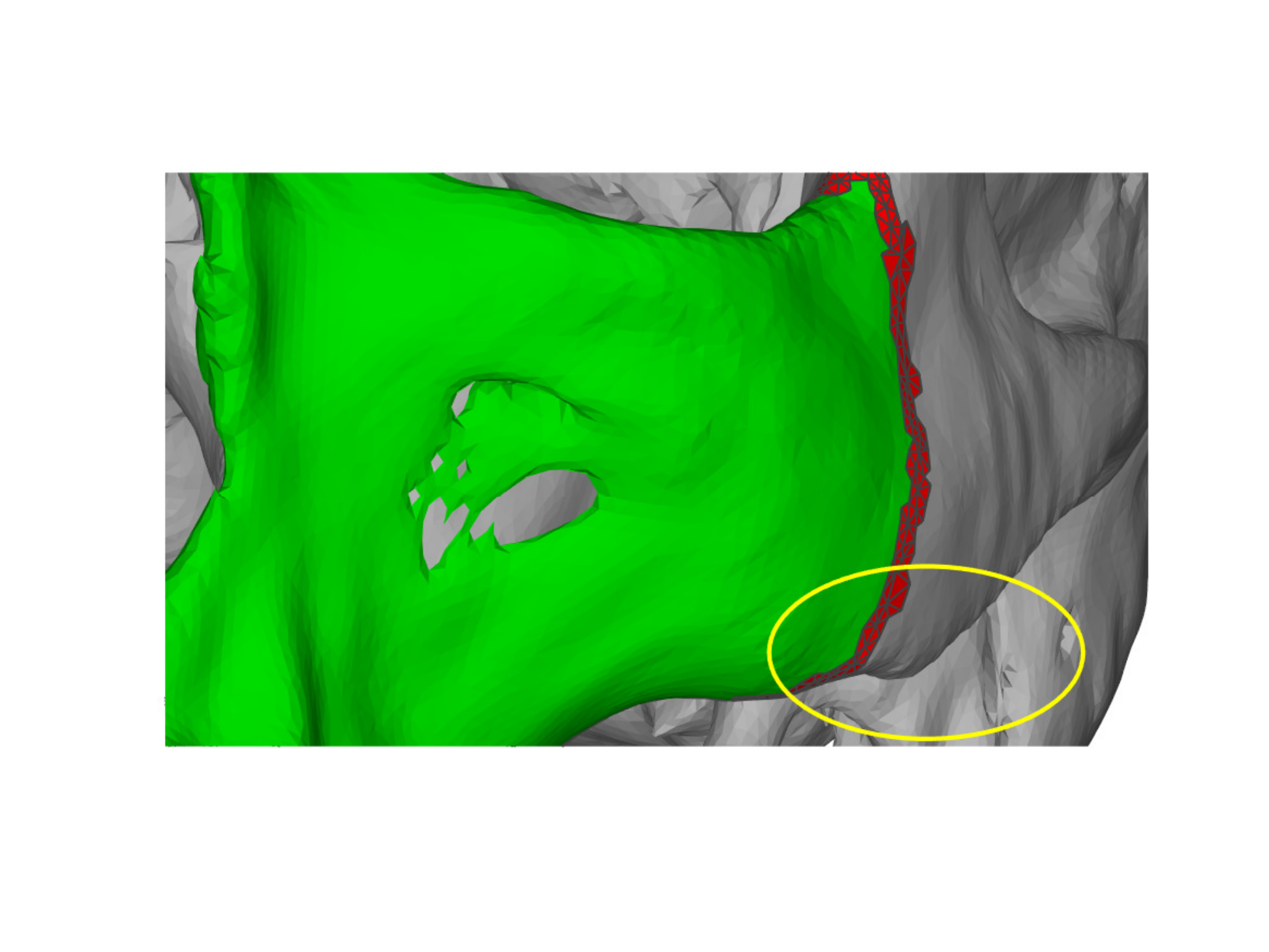}
                \caption{}
                \label{fig:skull_0522c0014_01_def_03_proj_tps_zoom}
        \end{subfigure}
        \caption{A skull extrapolation example for 20\% of the original surface removed.
        		     The known region is represented in gray, the unknown region is represented in green, and the boundary region is represented in red. (a)(d) PO extrapolation; the non-smooth transition is clearly visible. (b)(e) P+F, note the corruption of known vertex values by P+F in the gray region (yellow ellipse). (c)(f) P+TPS.} 
\label{fig:skull_extrap_examples}
\end{figure}
%%%%%%%%%%%%%%%%%%%%%%%%%%%%%%%%%%%%%%%%%%%%%%%%%%%%%%%%%%%%%
%%%%%%%%%%%%%%%%%%%%%%%%%%%%%%%%%%%%%%%%%%%%%%%%%%
\begin{figure}
        \centering
        \begin{subfigure}[b]{0.29\textwidth}
                \includegraphics[width=\textwidth, clip=true, trim=120 45 120 45]{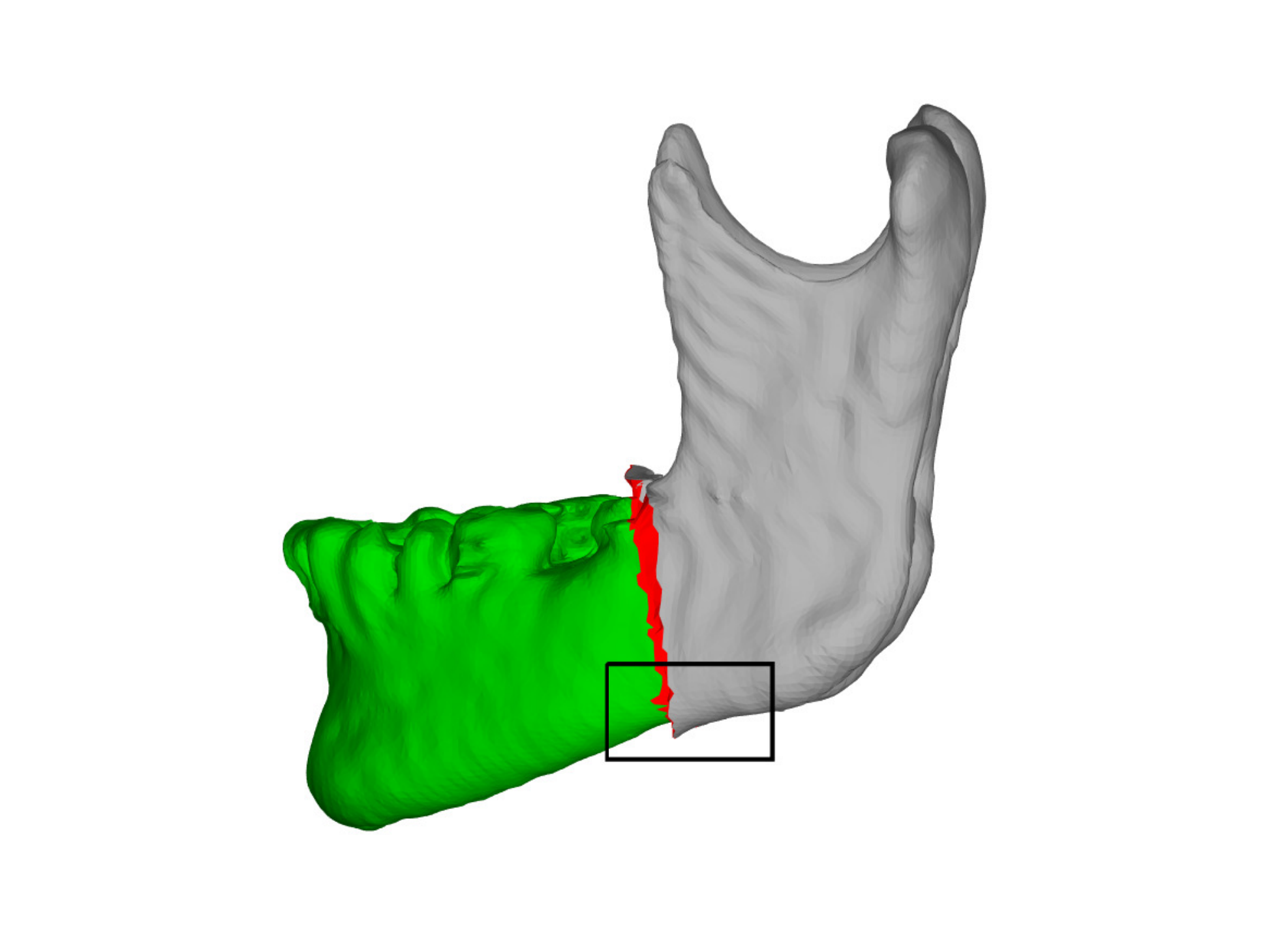}
                \caption{}
                \label{fig:mand_0522c0014_01_def_09_proj_only}
        \end{subfigure}
        \begin{subfigure}[b]{0.29\textwidth}
                \includegraphics[width=\textwidth, clip=true, trim=120 45 120 45]{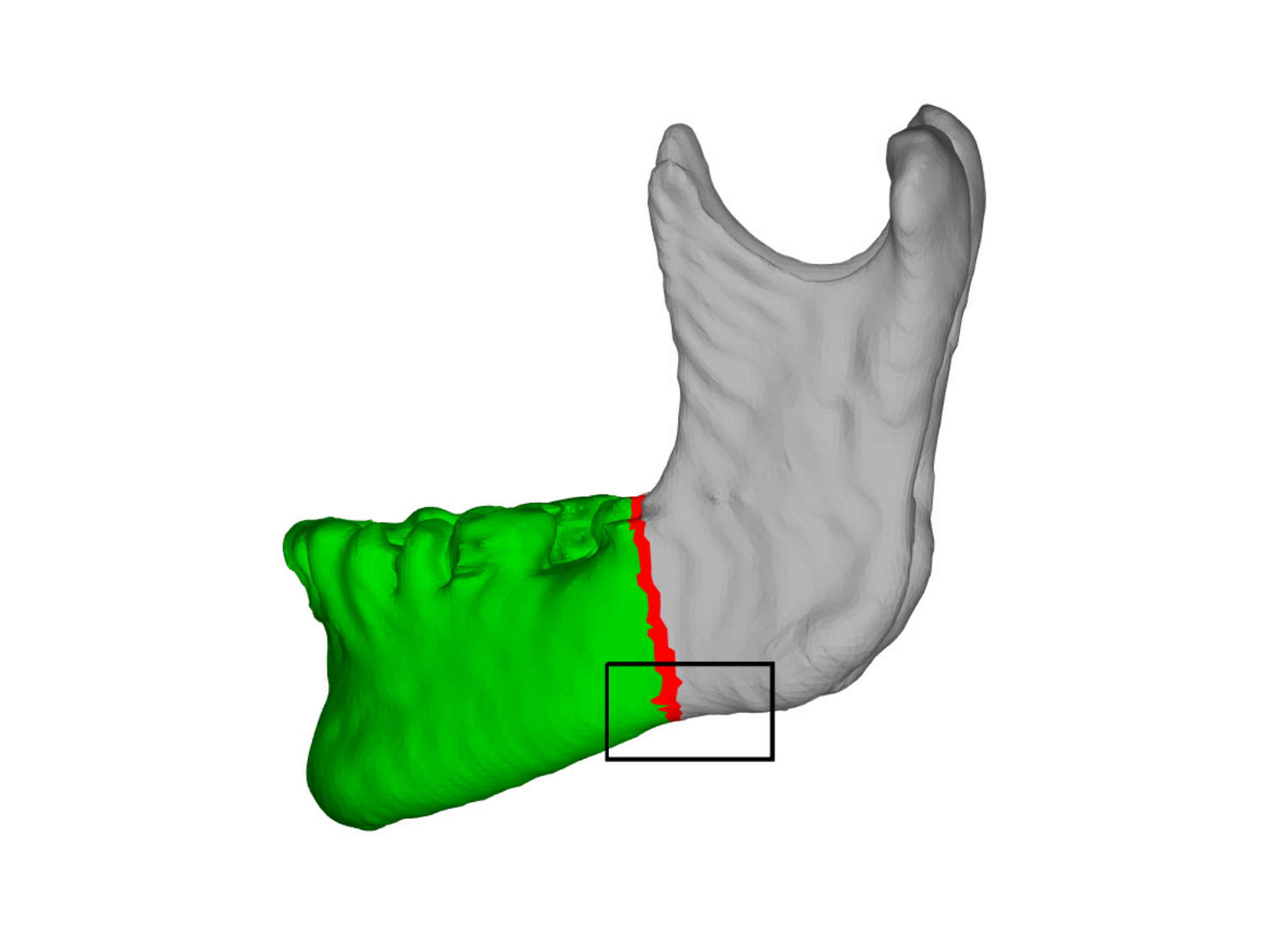}
                \caption{}
                \label{fig:mand_0522c0014_01_def_09_proj_feather}
        \end{subfigure}
        \begin{subfigure}[b]{0.29\textwidth}
                \includegraphics[width=\textwidth, clip=true, trim=120 45 120 45]{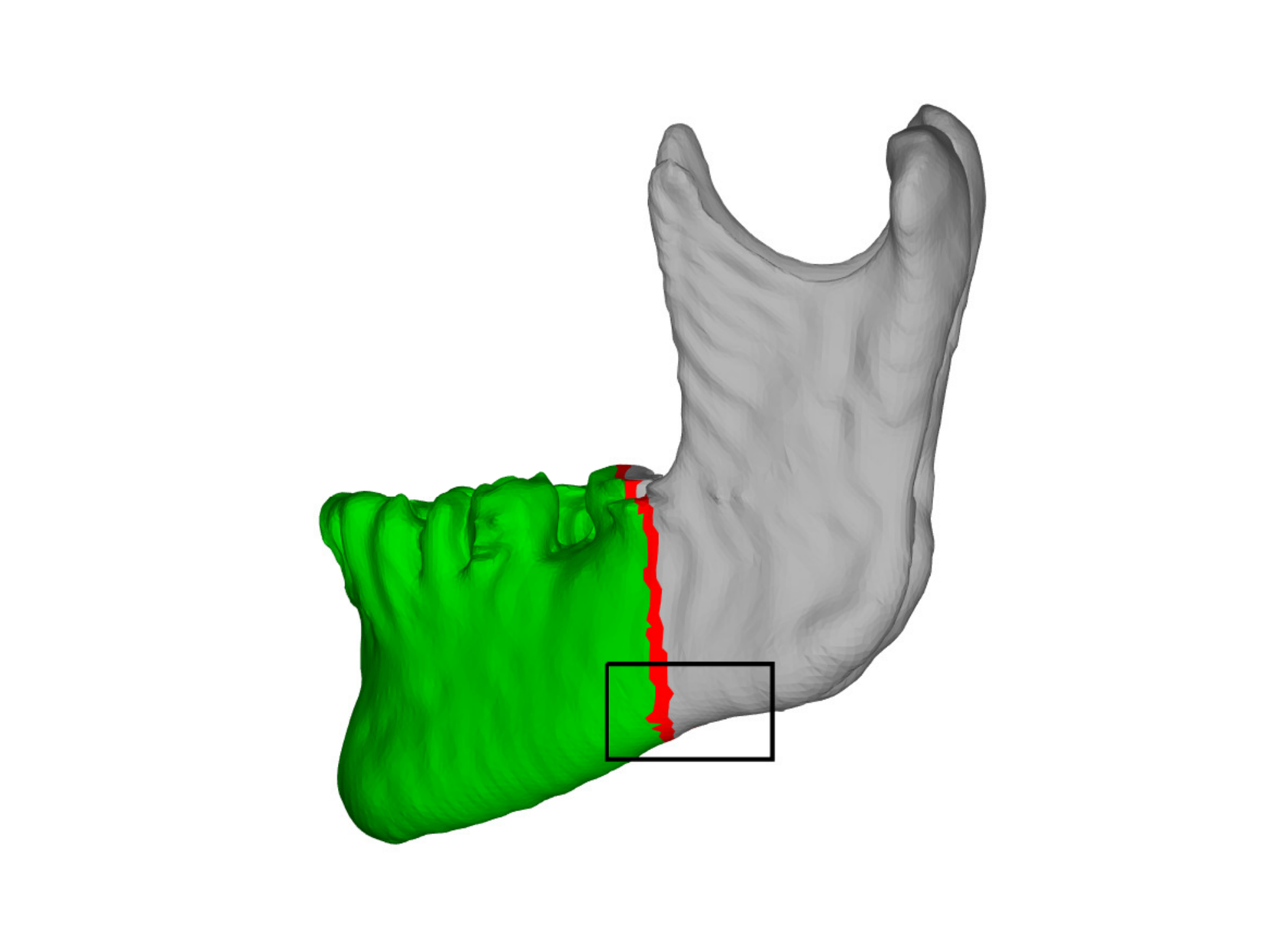}
                \caption{}
                \label{fig:mand_0522c0014_01_def_09_proj_tps}
        \end{subfigure}\\
        \begin{subfigure}[b]{0.29\textwidth}
                \includegraphics[width=\textwidth, clip=true, trim=130 120 100 45]{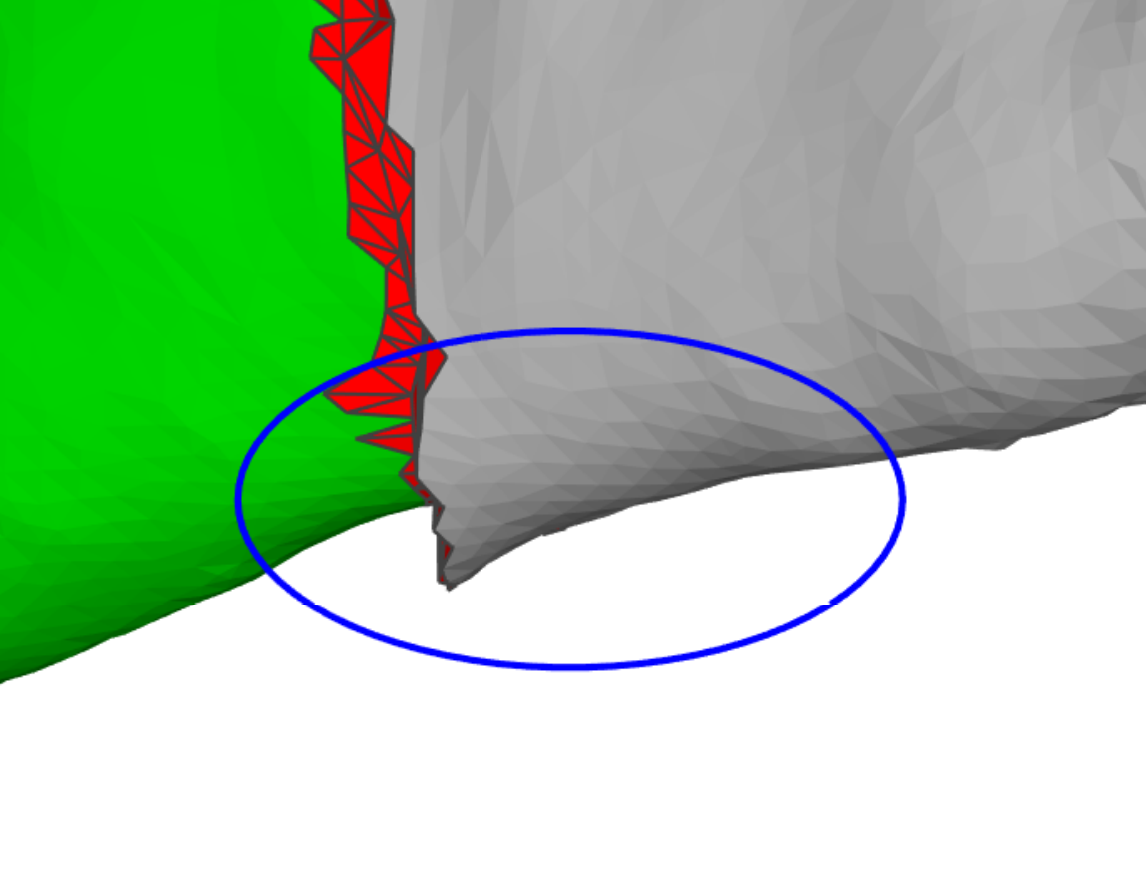}
                \caption{}
                \label{fig:mand_0522c0014_01_def_09_proj_only_zoom}
        \end{subfigure}
        \begin{subfigure}[b]{0.29\textwidth}
                \includegraphics[width=\textwidth, clip=true, trim=130 120 100 45]{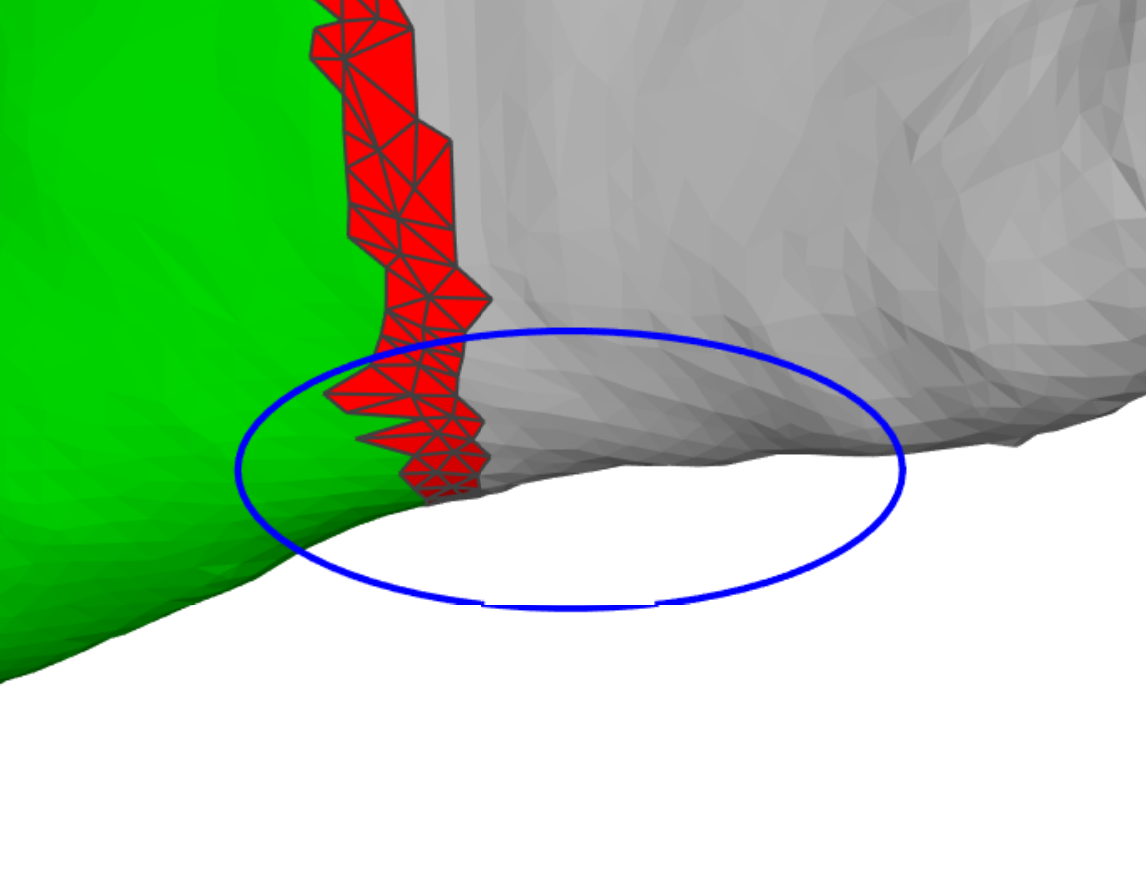}
                \caption{}
                \label{fig:mand_0522c0014_01_def_09_proj_feather_zoom}
        \end{subfigure}
        \begin{subfigure}[b]{0.29\textwidth}
                \includegraphics[width=\textwidth, clip=true, trim=130 120 100 45]{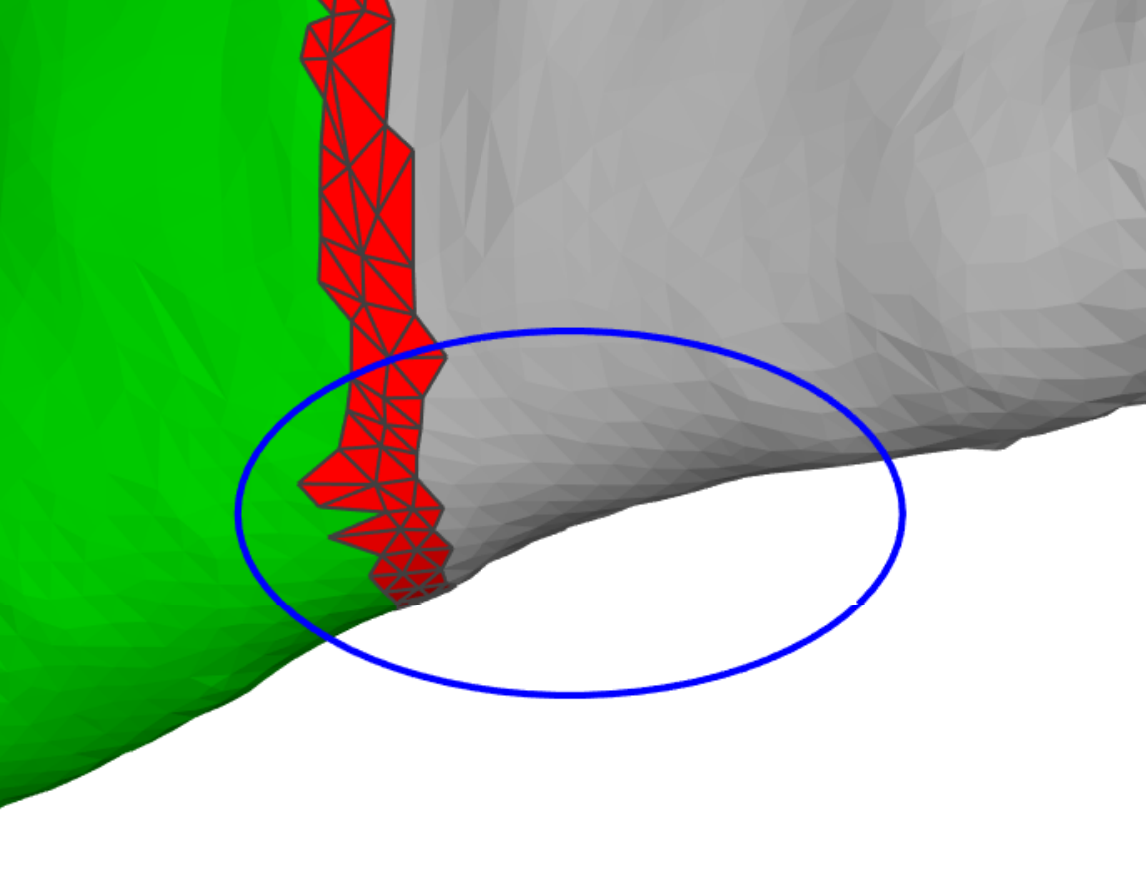}
                \caption{}
                \label{fig:mand_0522c0014_01_def_09_proj_tps_zoom}
        \end{subfigure}
        \caption{A mandible extrapolation example for 50\% of the original surface removed.
                     The known region is represented in gray, the unknown region is represented in green, and the boundary region is represented in red. (a)(d) PO extrapolation; note the non-smooth transition. (b)(e) P+F; note the corruption of known vertex values (blue ellipse). (c)(f) P+TPS.} 
\label{fig:mand_extrap_examples}
\end{figure}
%%%%%%%%%%%%%%%%%%%%%%%%%%%%%%%%%%%%%%%%%%%%%%%%%%%%%%%%%%%%%

For P+F, we used an overlap region with a maximum edge depth of 20; this was chosen to ensure a smooth transition.
For P+TPS, overlap regions with a maximum edge depth of 3 were used.
These values were chosen to reduce the computational burden of the TPS weight calculation, while still producing satisfactory results.
Figures \ref{fig:skull_extrap_examples} and \ref{fig:mand_extrap_examples} highlight specific examples of each extrapolation method for the skull and mandible, respectively.

\begin{figure}
        \centering
        \begin{subfigure}[b]{0.45\textwidth}
                \includegraphics[width=\textwidth]{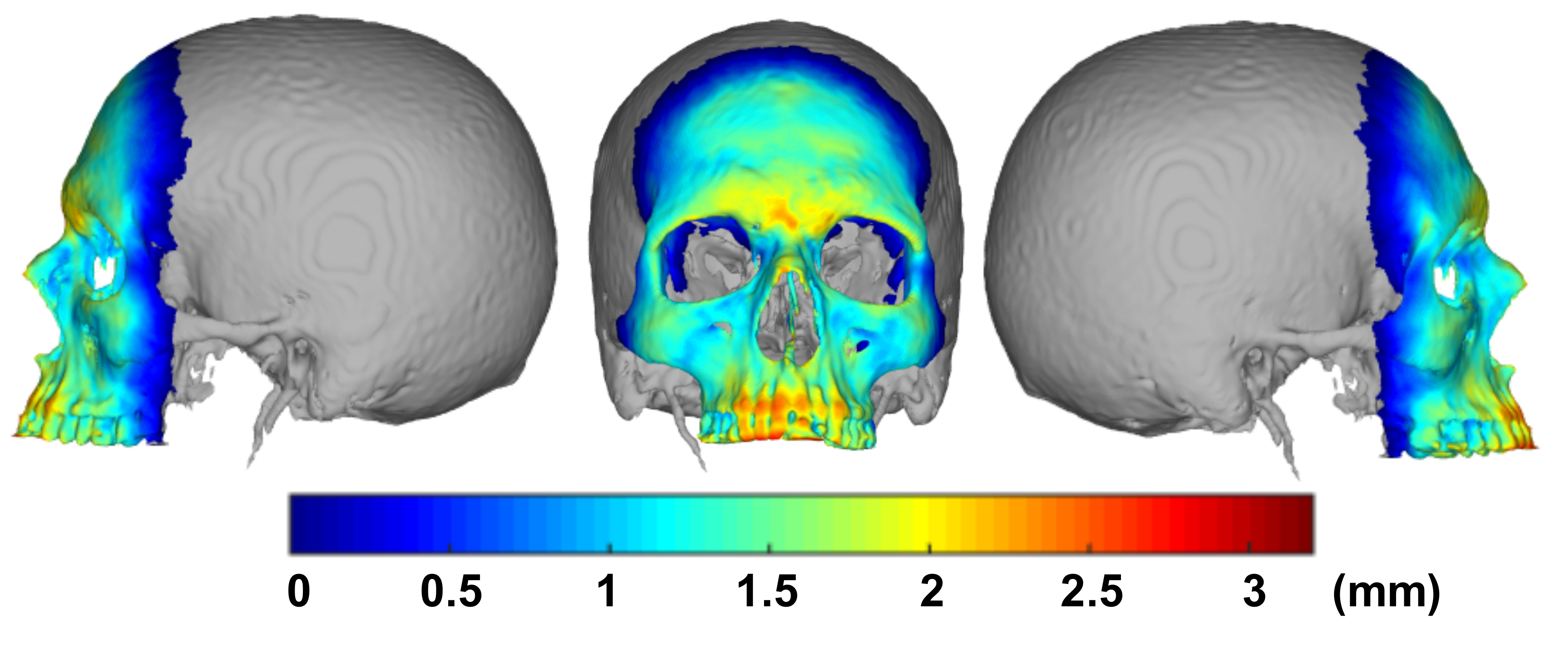}
                \caption{}
                \label{fig:skull_feather_extrap_heat}
        \end{subfigure}
        \begin{subfigure}[b]{0.45\textwidth}
                \includegraphics[width=\textwidth]{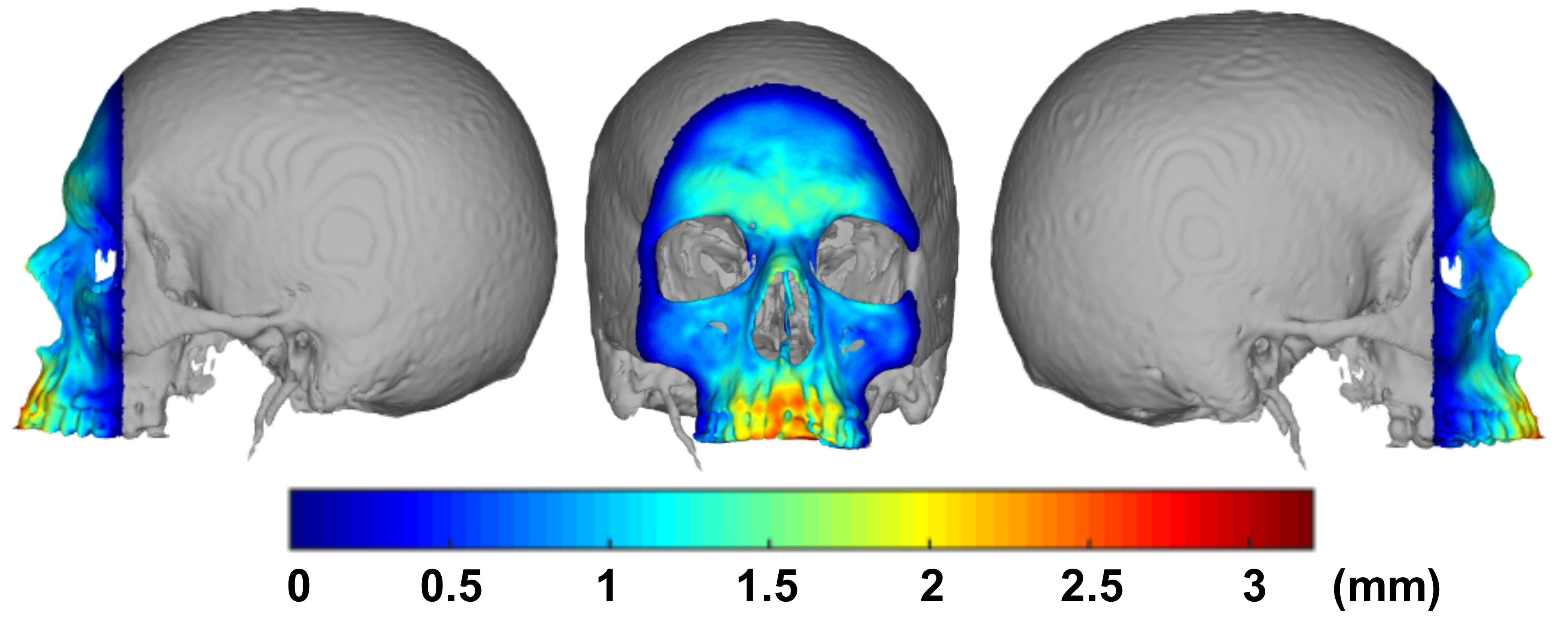}
                \caption{}
                \label{fig:skull_tps_extrap_heat}
        \end{subfigure}
        \caption{Skull heat maps of mean surface distances at each vertex generated over the course of the 20\% unknown anatomy leave-out iteration.
        		     (a) shows P+F and (b) shows P+TPS. Gray indicates a vertex value equal to the true surface. Overlay is the mean surface.
		     } \label{fig:skull_extrap_heat}
\end{figure}

\begin{figure}
        \centering
        \begin{subfigure}[b]{0.45\textwidth}
                \includegraphics[width=\textwidth]{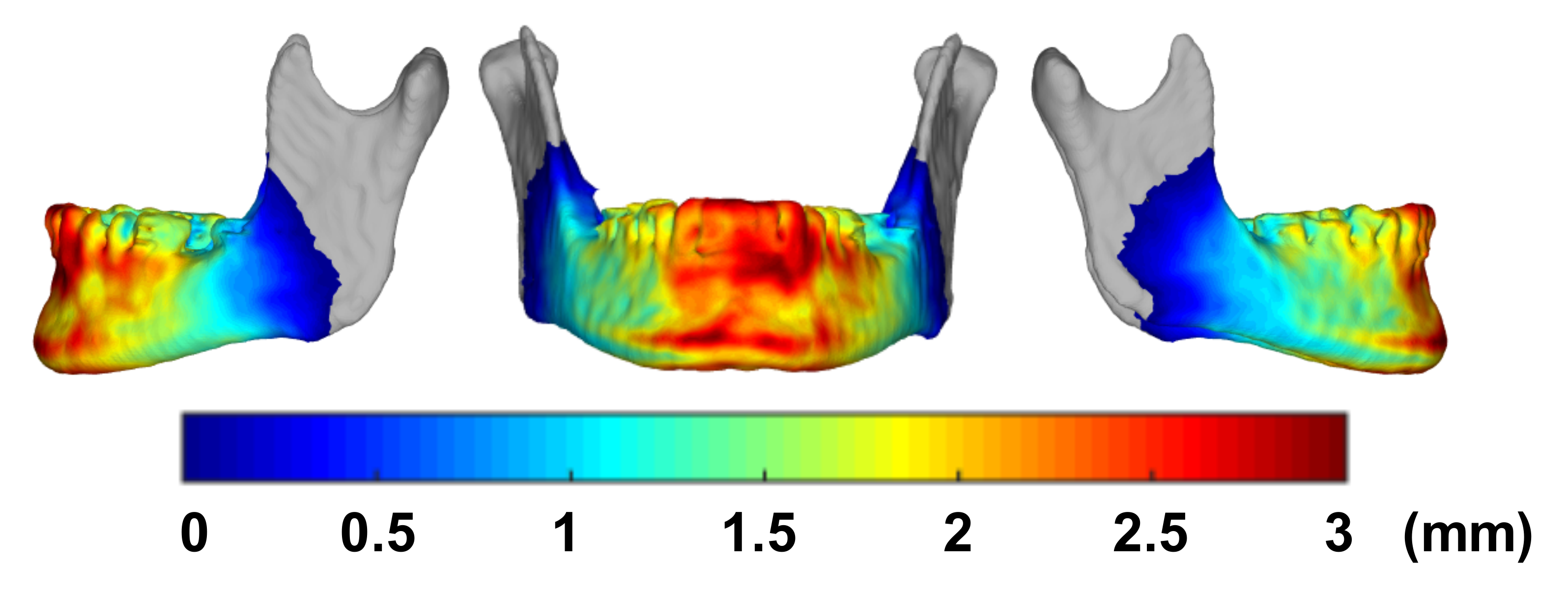}
                \caption{}
                \label{fig:mand_feather_extrap_heat}
        \end{subfigure}
        \begin{subfigure}[b]{0.45\textwidth}
                \includegraphics[width=\textwidth]{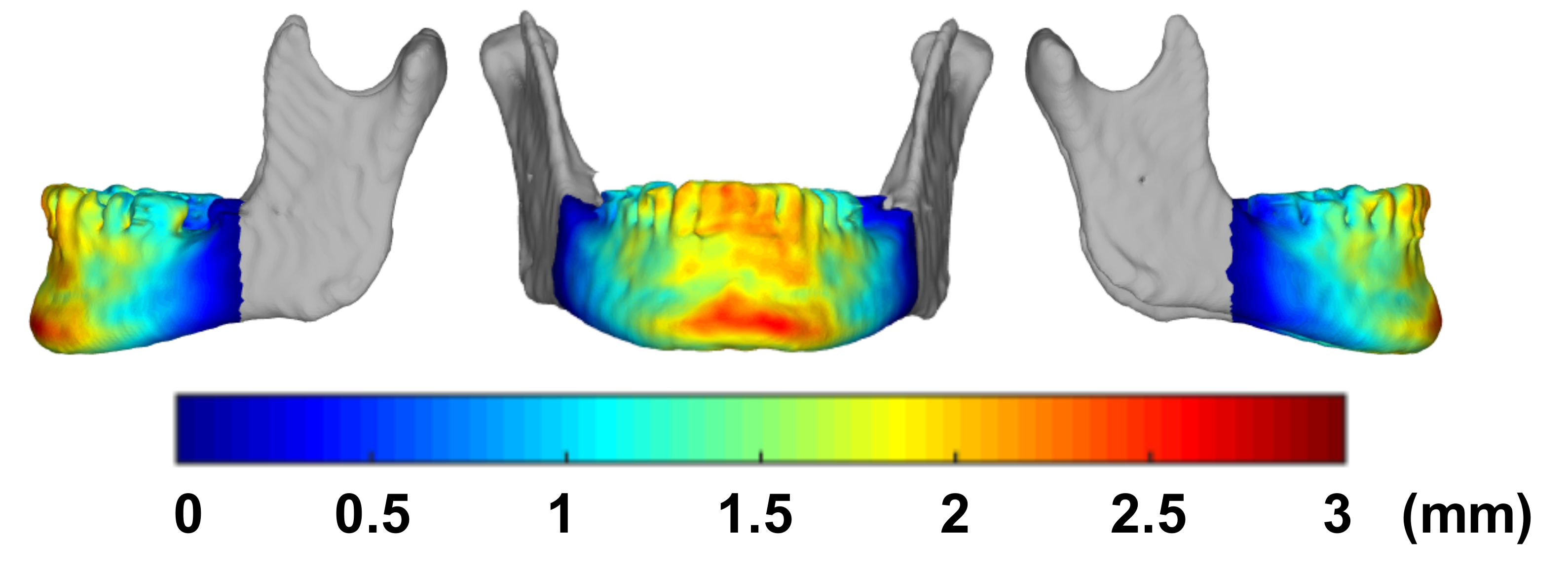}
                \caption{}
                \label{fig:mand_tps_extrap_heat}
        \end{subfigure}
        \caption{Mandible heat maps of mean surface distances at each vertex generated over the course of the 50\% unknown anatomy leave-out iteration.
        		     (a) shows P+F and (b) shows P+TPS. Gray indicates a vertex value equal to the true surface. Overlay is the mean surface.
		     } \label{fig:mand_extrap_heat}
\end{figure}

Figures \ref{fig:skull_extrap_heat} and \ref{fig:mand_extrap_heat} show the distribution of surface errors over the leave-out trials for specific cropping percentages.
The mean surface errors increase as the distance from known anatomy also increases, however figures \ref{fig:skull_extrap_examples} and \ref{fig:mand_extrap_examples} demonstrate that the performance of P+TPS degrades at a slower rate.

\begin{figure}
        \centering
        \begin{subfigure}[b]{0.3\textwidth}
                \includegraphics[width=\textwidth]{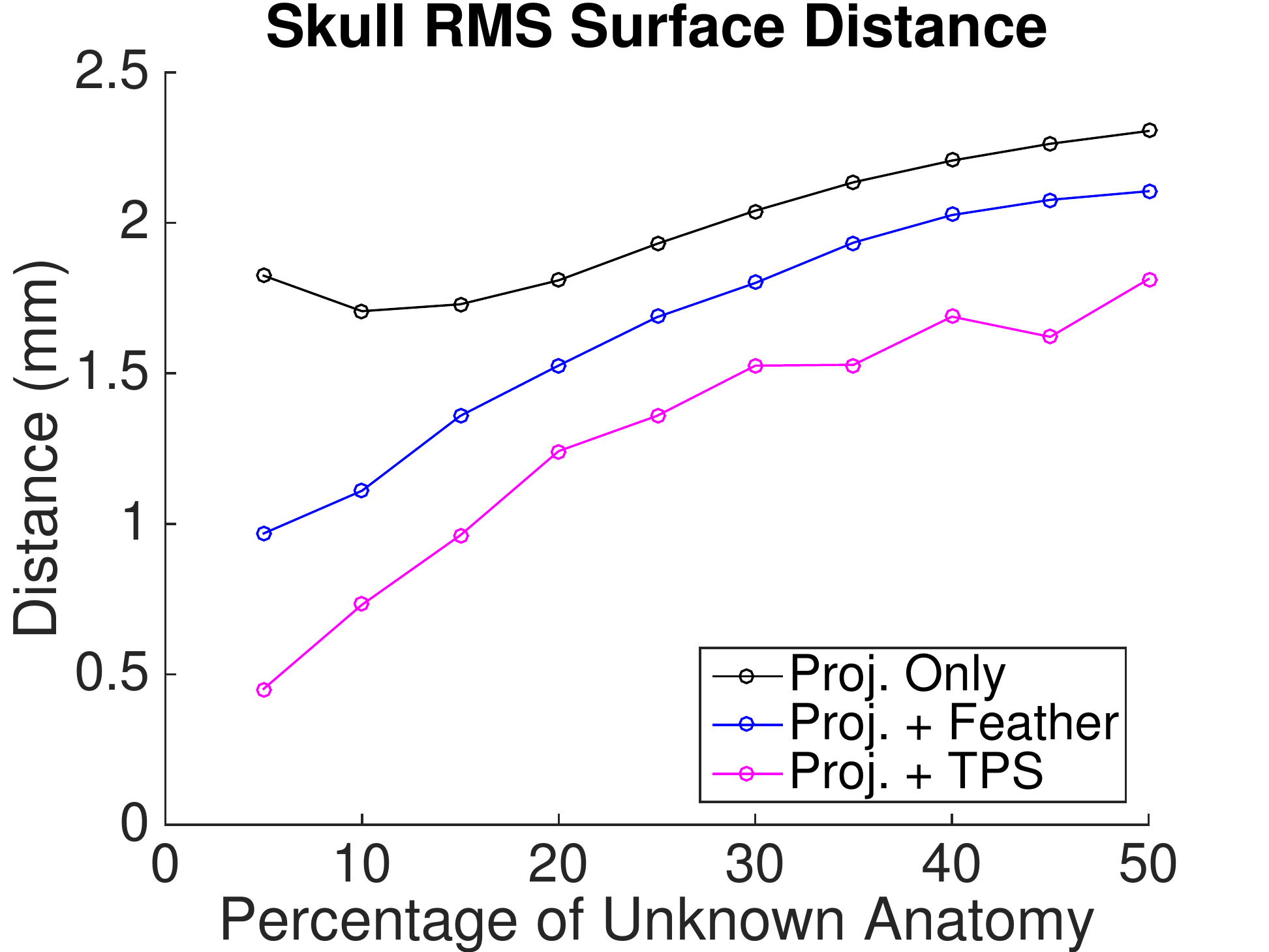}
                \caption{}
                \label{fig:skull_extrap_rms_sur}
        \end{subfigure}
        \begin{subfigure}[b]{0.3\textwidth}
                \includegraphics[width=\textwidth]{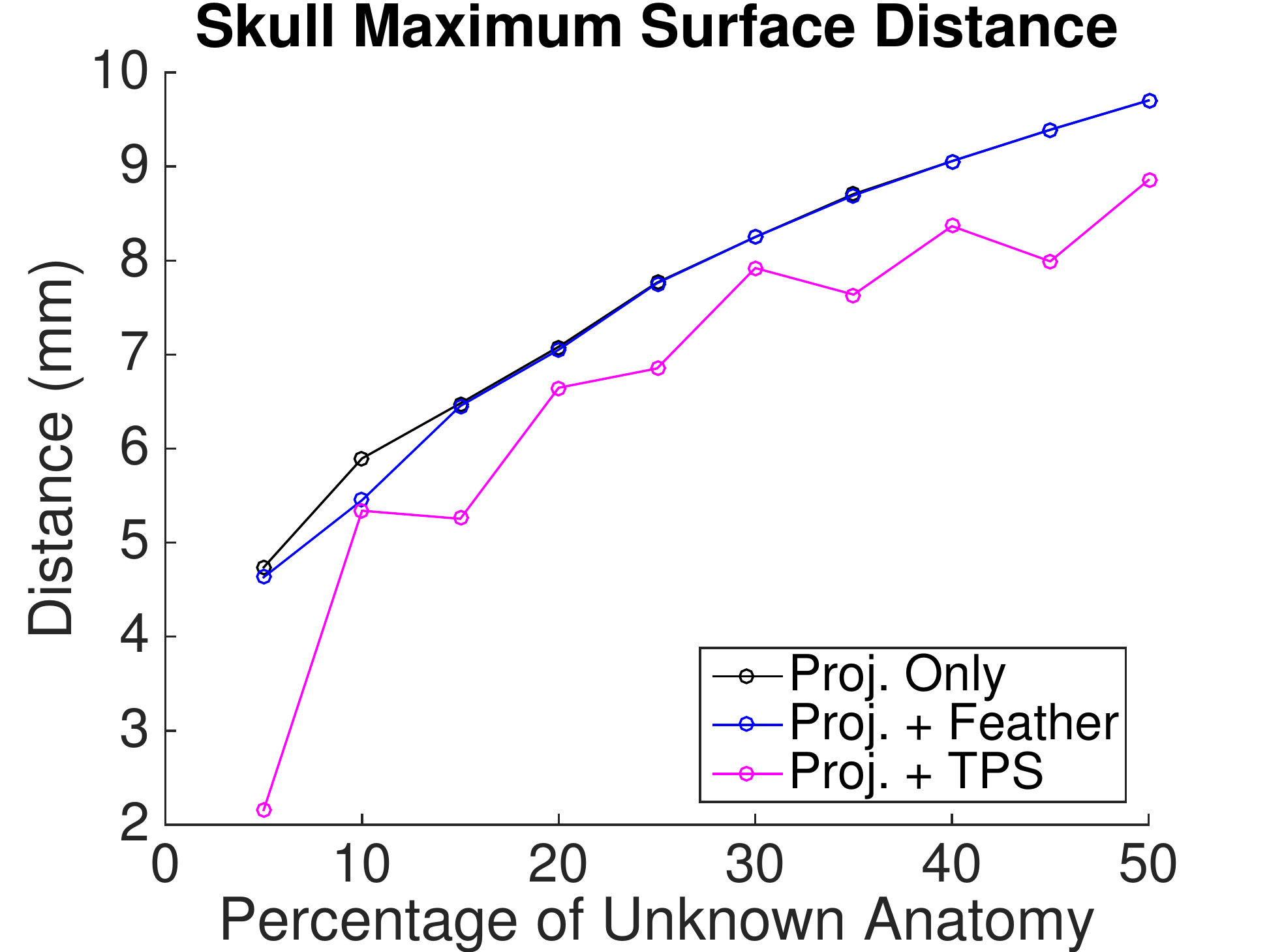}
                \caption{}
                \label{fig:skull_extrap_max_sur}
        \end{subfigure}
        \begin{subfigure}[b]{0.3\textwidth}
                \includegraphics[width=\textwidth]{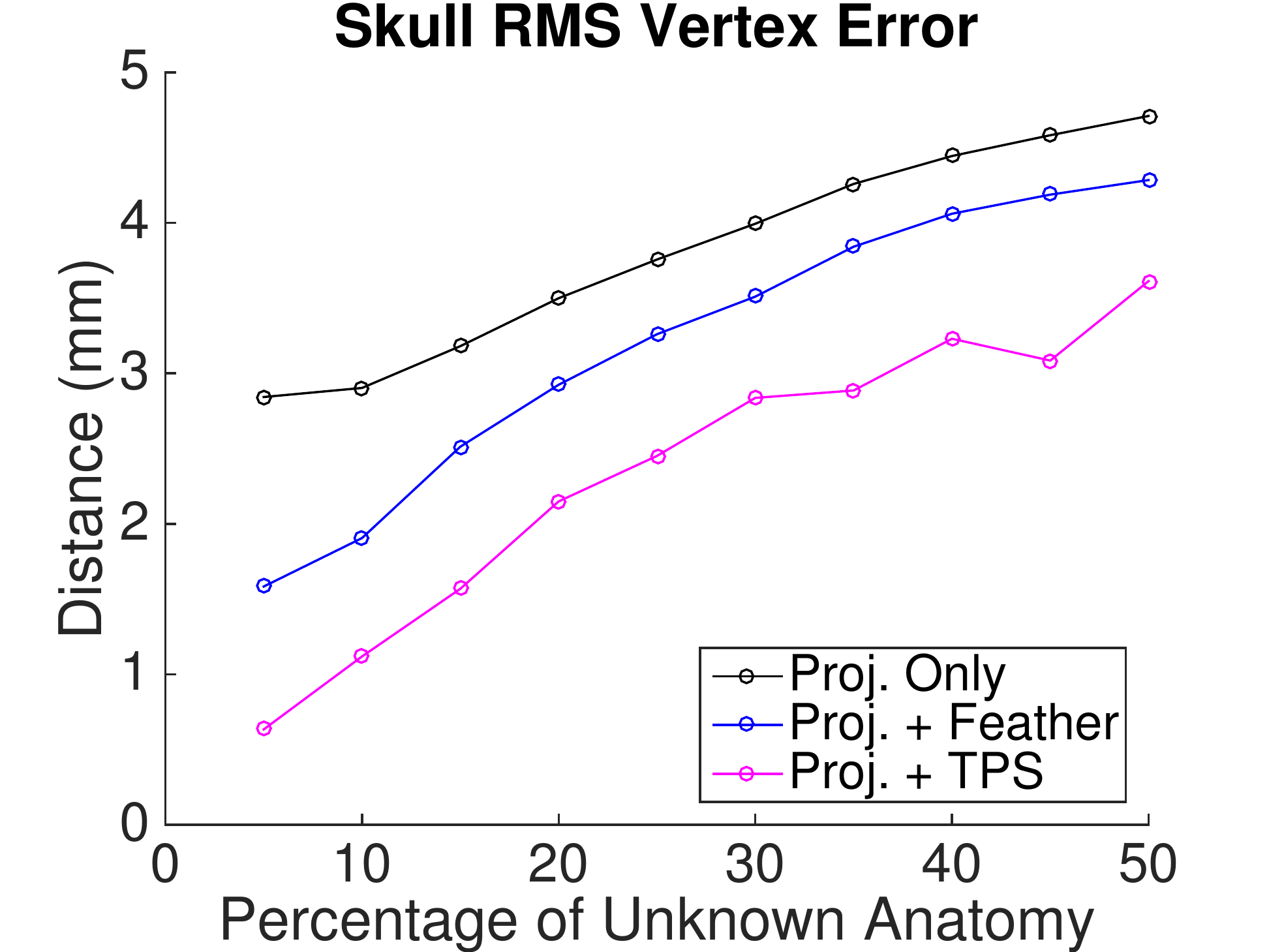}
                \caption{}
                \label{fig:skull_extrap_rms_vert}
        \end{subfigure}
        \caption{Statistics of the extrapolation leave-one-out tests for the skull surface.
        		     (a) RMS surface residuals.
		     (b) maximum surface residuals.
		     (c) RMS vertex displacement errors.
		     %Every mode of the SSM was used.
		     % when projecting the left-out-surface onto the SSM.
		     %The P+TPS exhibits smaller error measurements than P+F and PO.
		     %P+F results in smaller RMS error than PO, at the cost of modifying known vertex values.
		     } \label{fig:skull_extrap_dist_plots}
\end{figure}

\begin{figure}
        \centering
        \begin{subfigure}[b]{0.3\textwidth}
                \includegraphics[width=\textwidth]{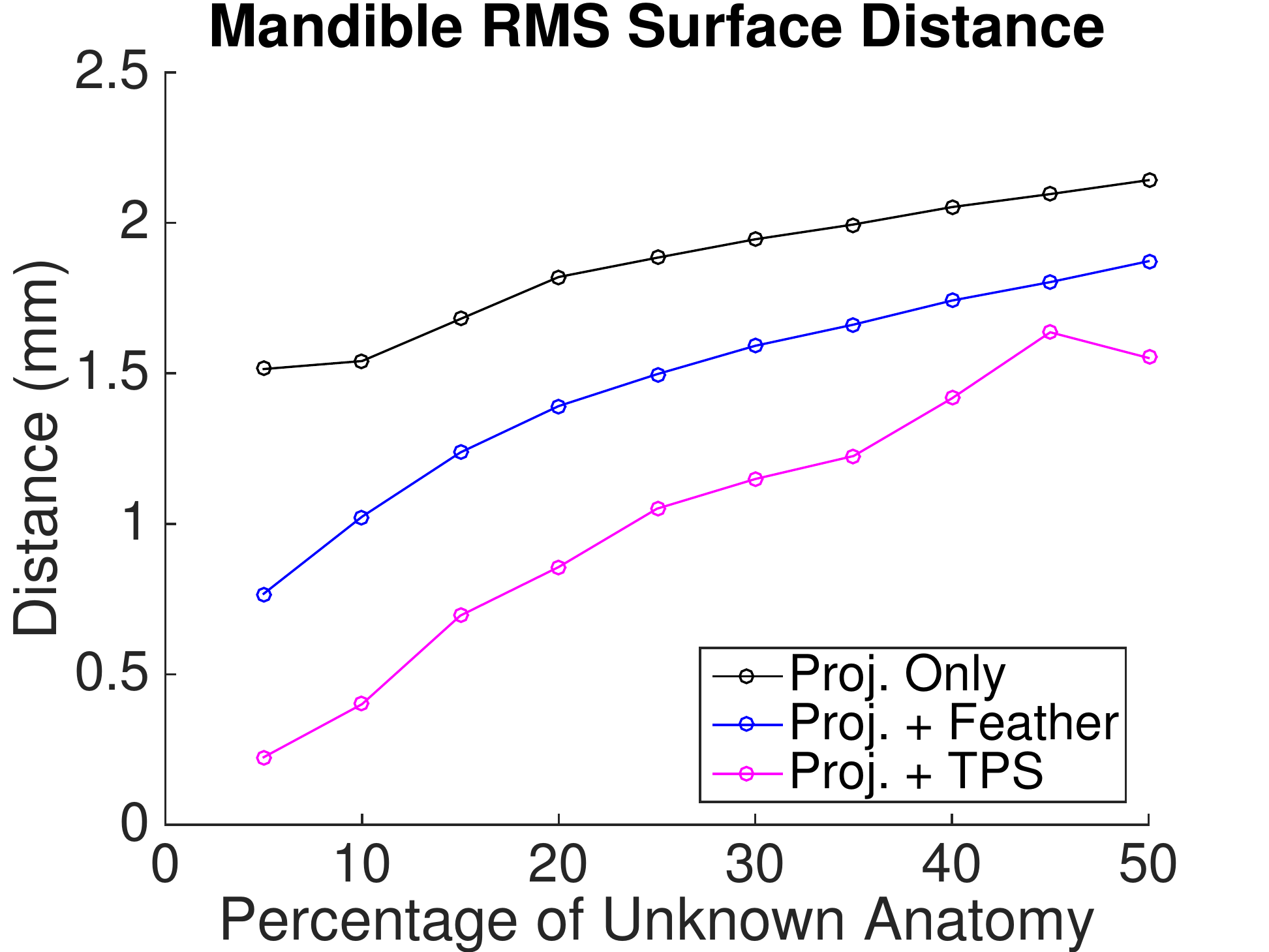}
                \caption{}
                \label{fig:mand_extrap_rms_sur}
        \end{subfigure}
        \begin{subfigure}[b]{0.3\textwidth}
                \includegraphics[width=\textwidth]{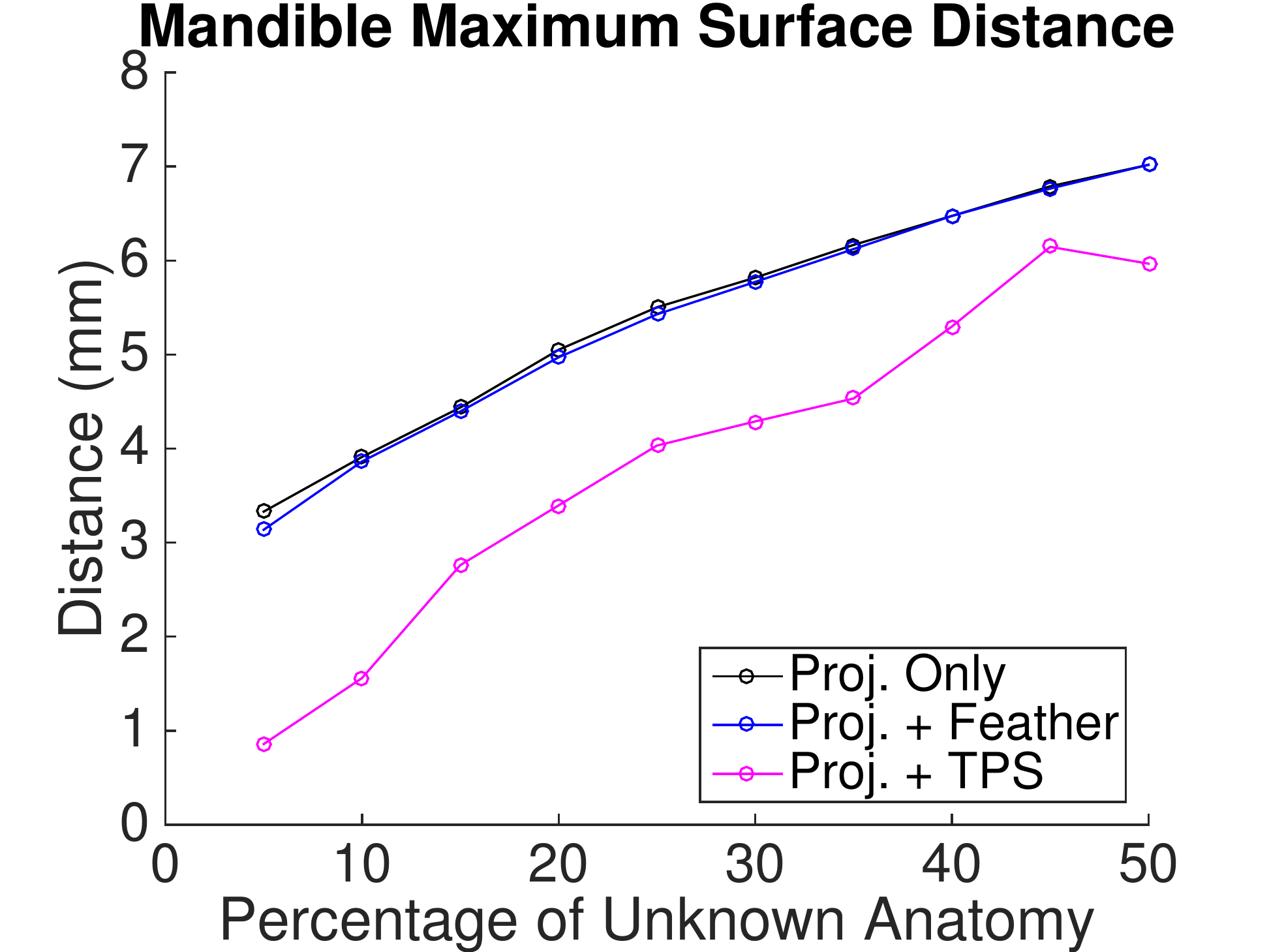}
                \caption{}
                \label{fig:mand_extrap_max_sur}
        \end{subfigure}
        \begin{subfigure}[b]{0.3\textwidth}
                \includegraphics[width=\textwidth]{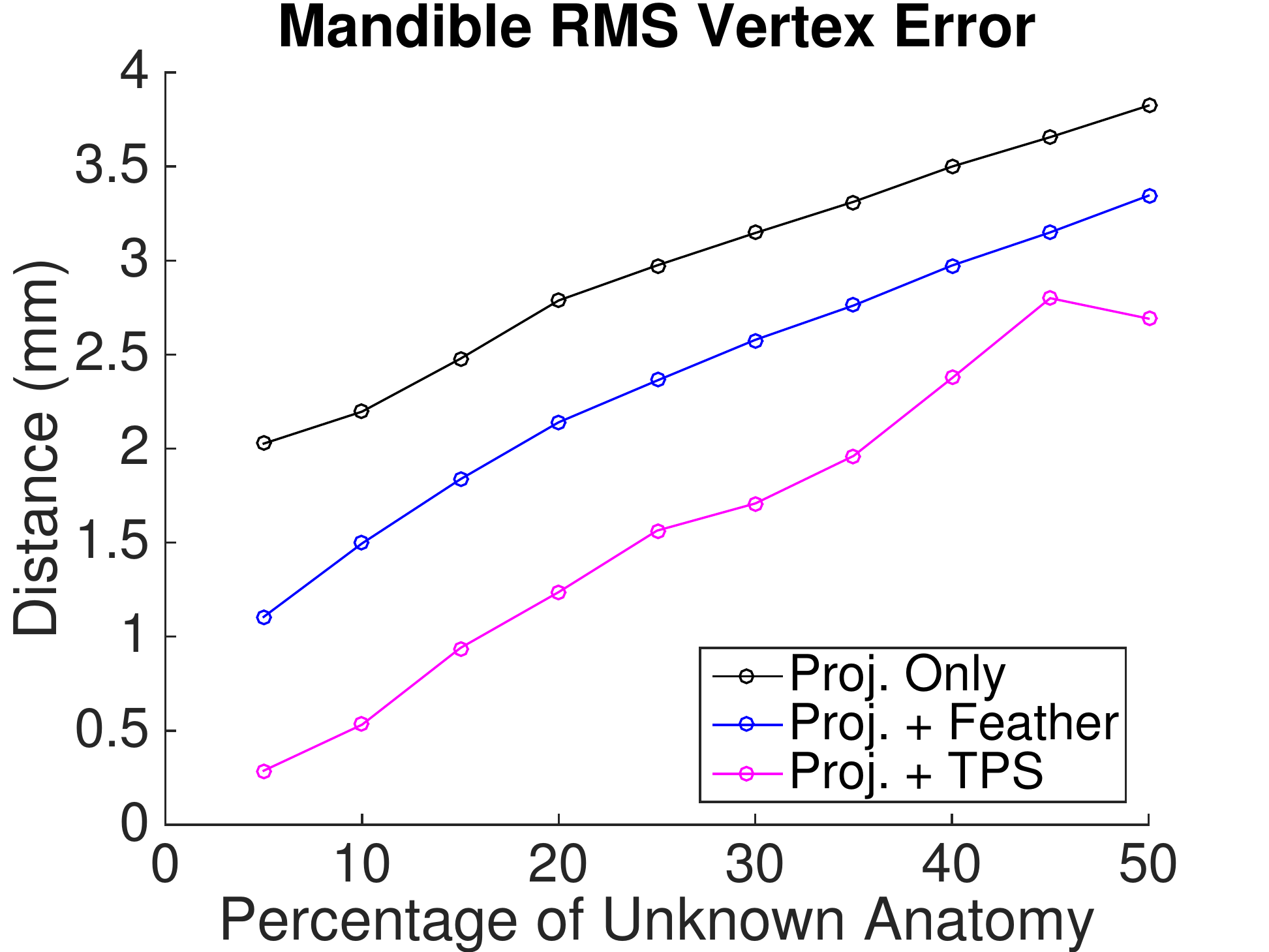}
                \caption{}
                \label{fig:mand_extrap_rms_vert}
        \end{subfigure}
        \caption{Statistics of the extrapolation leave-one-out tests for the mandible surface.
        		     (a) RMS surface residuals.
		     (b) maximum surface residuals.
		     (c) RMS vertex displacement errors.
		     %Every mode of the SSM was used.
		     % when projecting the left-out-surface onto the SSM.
		     %The P+TPS exhibits smaller error measurements than P+F and PO.
		     %P+F results in smaller RMS error than PO, at the cost of modifying known vertex values.
		     } \label{fig:mand_extrap_dist_plots}
\end{figure}

Figures \ref{fig:skull_extrap_dist_plots} and \ref{fig:mand_extrap_dist_plots} show error statistics of the extrapolated regions for each cropping size over the entire set of leave-out trials.
For the skull, P+TPS showed: an average RMS surface error improvement of 0.36 mm and 0.70 mm over P+F and PO, an average maximum surface error improvement of 0.94 mm and 1.00 mm over P+F and PO, and an average RMS vertex error improvement of 0.85 mm and 1.46 mm over P+F and PO.
For the mandible, P+TPS showed: an average RMS surface error improvement of 0.43 mm and 0.84 mm over P+F and PO, an average maximum surface error improvement of 1.51 mm and 1.56 mm over P+F and PO, and an average RMS vertex error improvement of 0.76 mm and 1.38 mm over P+F and PO.

\begin{figure}
        \centering
        \includegraphics[width=0.6\textwidth]{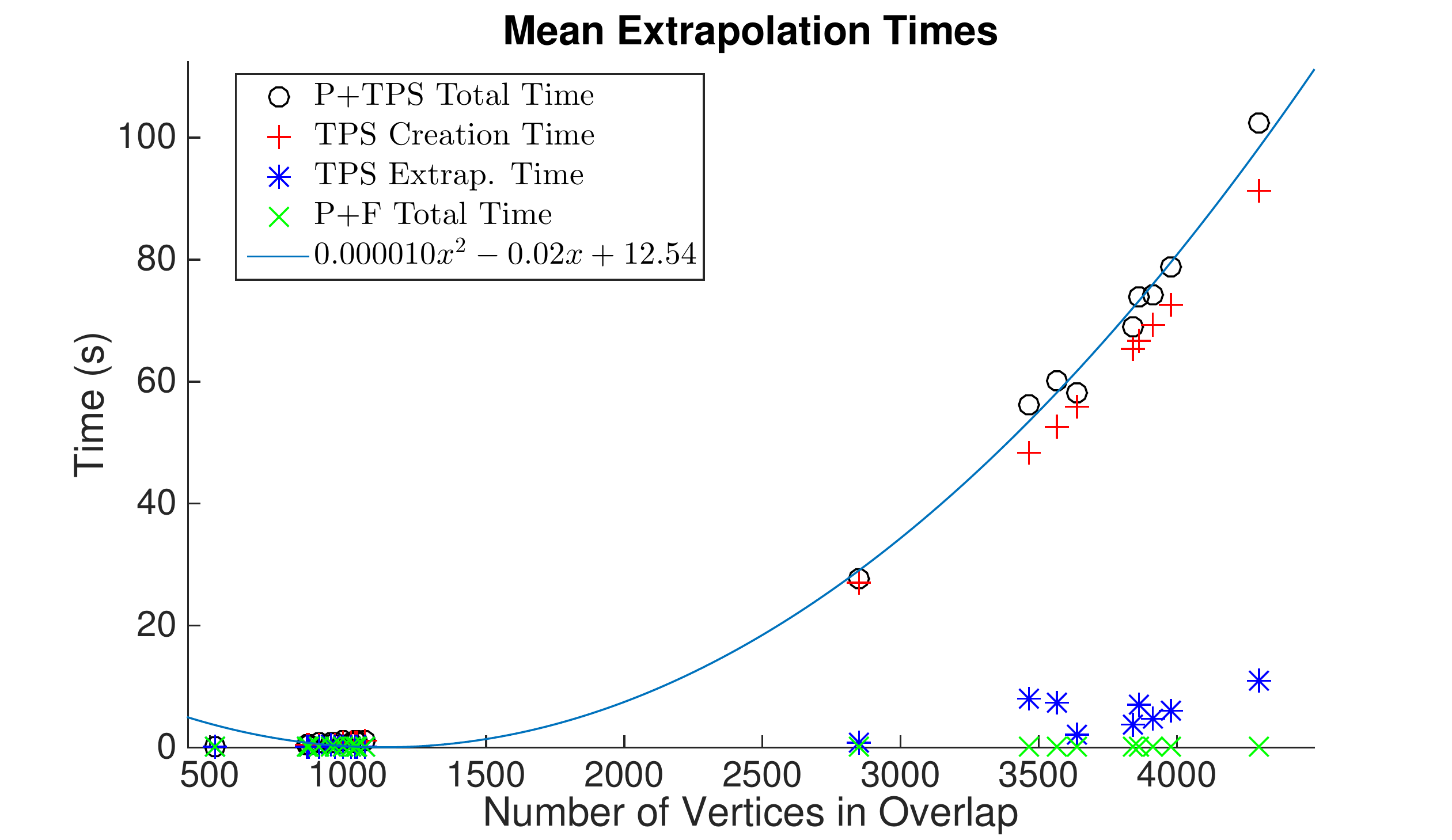}
        \caption{A plot of mean recorded runtimes for P+F, P+TPS, the TPS creation portion of P+TPS, and the TPS evaluation portion of P+TPS.
        		    A Quadratic least-squares regression is shown for the P+TPS runtimes.}
        \label{fig:tps_runtimes}
\end{figure}

As previously mentioned, P+TPS is significantly slower than P+F.
During the leave-out analysis, execution times were recorded for each extrapolation method.
All executions were executed on a machine with a single Intel Xeon E5-2640 running at 2.5 GHz with 6 physical cores, but allowing for 12 virtual cores via hyper-threading.
The PO and P+F extrapolation methods were performed in a single thread.
The TPS creation was run with 3 threads due to the limitation of the single-threaded QR factorization and the TPS evaluation was run with 12 threads.
Figure \ref{fig:tps_runtimes} shows the mean runtimes for P+F and P+TPS as a function of the number of vertices in the overlap region.
The largest recorded PO runtime was 3.5 ms and the largest recorded runtime for P+F was 4.5 ms, so it is sufficient to only plot P+F in comparison to P+TPS.
The execution time of the TPS evaluation portion of P+TPS is a function of the number of vertices in the unknown region and the number of vertices in the overlap region, however it is useful to see from figure \ref{fig:tps_runtimes} that the TPS creation portion dominates the total percentage of runtime.
%%%%%%%%%%%%%%%%%%%%%%%%%%%%%%%%%%%%%%%%%%%%%%%%%%
%%%%%%%%%%%%%%%%%%%%%%%%%%%%%%%%%%%%%%%%%%%%%%%%%%
%%%%%%%%%%%%%%%%%%%%%%%%%%%%%%%%%%%%%%%%%%%%%%%%%%
%%%%%%%%%%%%%%%%%%%%%%%%%%%%%%%%%%%%%%%%%%%%%%%%%%
%%%%%%%%%%%%%%%%%%%%%%%%%%%%%%%%%%%%%%%%%%%%%%%%%%%%%%%%%%%%%
%%%%%%%%%%%%%%%%%%%%%%%%%%%%%%%%%%%%%%%%%%%%%%%%%%
%%%%%%%%%%%%%%%%%%%%%%%%%%%%%%%%%%%%%%%%%%%%%%%%%%%%%%%%%%%%%
%%%%%%%%%%%%%%%%%%%%%%%%%%%%%%%%%
%%%%%%%%%%%%%%%%%%%%%%%%%%%%%%%%%
\section{CONCLUSIONS}
\label{sec:conc}  % \label{} allows reference to this section
We have demonstrated that smooth extrapolation of anatomy is possible via a SSM and a TPS. P+TPS outperformed, both PO and P+F, while still preserving known regions of anatomy. Further work is required to perform extrapolation on a mesh with arbitrary topology; this will require deformable registration to the SSM and ``topology transfer.'' P+TPS is significantly more computationally demanding than the P+F, therefore some investigation into approximate TPS methodologies needs to be conducted in order to perform an analysis on the optimal size of overlap regions to be used.
With advancements in smooth extrapolation of unknown anatomy, the potential benefits for computer-assisted surgery will continue to expand, and be most applicable to those fields required to manipulate complex skeletal components including craniomaxillofacial and orthopedic surgery.
%%%%%%%%%%%%%%%%%%%%%%%%%%%%%%%%%%%%%%%%%%%%%%%%%%%%%%%%%%%%%
\acknowledgments     %>>>> equivalent to \section*{ACKNOWLEDGMENTS}
%   NIH-1R01EB016703 is Osteolysis
% NIH-R01EB006839 PAO
This research is supported by NIH-1R01EB016703, NIH-R01EB006839, JHU internal funds, Academic Scholar Award (American Association of Plastic Surgeons),
2012 ATIP Award (Johns Hopkins Institute of Clinical and Translational Research),
and the 2013 Abell Foundation Prize (Johns Hopkins Alliance for Science and Technology).
%%%%%%%%%%%%%%%%%%%%%%%%%%%%%%%%%%%%%%%%%%%%%%%%%%%%%%%%%%%%%
%%%%% References %%%%%
\bibliography{references}   %>>>> bibliography data in report.bib
\bibliographystyle{spiebib}   %>>>> makes bibtex use spiebib.bst
\end{document}